%% file: iclr2027_conference.tex
\PassOptionsToPackage{table}{xcolor}
\documentclass{article} 
\usepackage{iclr2027_conference,times}

\input{math_commands.tex}

\usepackage{hyperref}
\usepackage{url}
\usepackage{cleveref}
\usepackage{graphicx}
\usepackage{array}
\usepackage{booktabs}
\usepackage{multirow}
\usepackage{wrapfig}
\usepackage{placeins}
\usepackage{algorithm}
\usepackage{algpseudocode}
\usepackage{caption}

\newcommand{\ours}{PixGR}

\newcommand{\eg}{e.g.}
\newcommand{\best}[1]{\textbf{#1}}
\newcommand{\secondbest}[1]{\underline{#1}}
\definecolor{ourscell}{HTML}{E4F2FC}

\crefname{section}{Sec.}{Secs.}
\Crefname{section}{Sec.}{Secs.}
\crefname{equation}{Eq.}{Eqs.}
\Crefname{equation}{Eq.}{Eqs.}
\crefname{figure}{Fig.}{Figs.}
\Crefname{figure}{Fig.}{Figs.}
\crefname{table}{Tab.}{Tabs.}
\Crefname{table}{Tab.}{Tabs.}

\title{PixGR: Pixel-Space Generative Decoding for Fast Feed-Forward 3D Reconstruction}

\author{
Ranran Huang$^{1}$ \quad
Wenjing Bian$^{2}$ \quad
Weixun Luo$^{1}$ \quad
Ye Mao$^{1}$ \quad
Krystian Mikolajczyk$^{1}$
\\[0.4em]
{\normalfont
$^{1}$Imperial College London
\qquad
$^{2}$University of Tübingen
}
}

\iclrfinalcopy 
\begin{document}

\maketitle
\lhead{Preprint}

\begin{abstract}
\input{secs/abs}
\end{abstract}

\input{secs/intro}
\input{secs/related}
\input{secs/preliminary}
\input{secs/methods}
\input{secs/experiments}
\input{secs/conclusion}

\bibliography{reference}
\bibliographystyle{iclr2027_conference}
\newpage
\appendix
\input{secs/appendix}

\end{document}

%% file: math_commands.tex
\usepackage{amsmath,amsfonts,bm}

\def\eqref#1{equation~\ref{#1}}

\def\1{\bm{1}}

\def\ra{{\textnormal{a}}}

\def\rx{{\textnormal{x}}}

\def\rva{{\mathbf{a}}}

\def\erva{{\textnormal{a}}}

\def\ervx{{\textnormal{x}}}

\def\rmA{{\mathbf{A}}}

\def\vmu{{\bm{\mu}}}
\def\vtheta{{\bm{\theta}}}
\def\va{{\bm{a}}}

\def\ve{{\bm{\epsilon}}}

\def\vv{{\bm{v}}}

\def\vx{{\bm{x}}}

\def\vz{{\bm{z}}}

\def\eva{{a}}

\def\mA{{\bm{A}}}

\def\mH{{\bm{H}}}
\def\mI{{\bm{I}}}
\def\mJ{{\bm{J}}}

\def\mX{{\bm{X}}}

\def\mSigma{{\bm{\Sigma}}}

\DeclareMathAlphabet{\mathsfit}{\encodingdefault}{\sfdefault}{m}{sl}
\SetMathAlphabet{\mathsfit}{bold}{\encodingdefault}{\sfdefault}{bx}{n}
\newcommand{\tens}[1]{\bm{\mathsfit{#1}}}
\def\tA{{\tens{A}}}

\def\tX{{\tens{X}}}

\def\gG{{\mathcal{G}}}

\def\sA{{\mathbb{A}}}
\def\sB{{\mathbb{B}}}
\def\sC{{\mathbb{C}}}

\def\sN{{\mathbb{N}}}

\def\sS{{\mathbb{S}}}
\def\sT{{\mathbb{T}}}

\def\emA{{A}}

\newcommand{\etens}[1]{\mathsfit{#1}}

\def\etA{{\etens{A}}}

\newcommand{\E}{\mathbb{E}}

\newcommand{\R}{\mathbb{R}}

\newcommand{\KL}{D_{\mathrm{KL}}}
\newcommand{\Var}{\mathrm{Var}}

\newcommand{\Cov}{\mathrm{Cov}}

\newcommand{\normltwo}{L^2}
\newcommand{\normlp}{L^p}

\newcommand{\parents}{Pa} 

\newcommand{\e}{\bm{\epsilon}}

\renewcommand{\sC}{{\text{cxt}}}
\renewcommand{\sT}{{\text{tgt}}}
\renewcommand{\sN}{{\text{nov}}}

%% file: secs/abs.tex
We present \ours{}, an efficient framework that integrates lightweight pixel-space generative modeling into feed-forward 3D Gaussian reconstruction to model scene uncertainty. Context observations are encoded once into a scene latent representation, which conditions generative decoding from noisy target views to produce a shared 3D Gaussian representation. We formulate this process using pixel-space flow matching, enabling end-to-end optimization of 3D reconstruction with direct pixel-space generative supervision through differentiable rendering. This avoids VAE-based latent compression and error propagation inherent in previous two-stage pipelines. To preserve feed-forward efficiency, \ours{} caches the scene representation and performs iterative inference using a lightweight DiT decoder. Experiments demonstrate that \ours{} outperforms regression-based approaches in challenging sparse-view settings, particularly with widely spaced observations and view extrapolation. Meanwhile, it maintains sub-second inference and remains competitive with substantially heavier pretrained generative models. Our project page is available at \url{https://ranrhuang.github.io/pixgr/}.

%% file: secs/intro.tex
\section{Introduction}
\label{sec:introduction}


Reconstructing 3D scenes from a few images is a long-standing and fundamental problem in computer vision. Recent feed-forward generalizable methods have made remarkable progress by directly regressing 3D scene representations from input observations~\citep{charatan2024pixelsplat,chen2024mvsplat,huang2025spfsplat,xu2025depthsplat,ye2025noposplat,ye2026yonosplat}, as shown in \cref{fig:pipeline_comparison}a. These regression-based methods offer efficient single-pass inference and achieve impressive reconstruction quality in well-observed regions. However, under sparse-view observations, the underlying geometry and appearance of a 3D scene are weakly constrained, admitting multiple possible 3D explanations and introducing substantial reconstruction uncertainty. A deterministic regressor cannot explicitly model such ambiguity and instead commits to an averaged estimate, often resulting in blurry reconstructions or failures in occluded and unseen regions.

To address the high ambiguity of sparse-view reconstruction, generative modeling is introduced due to its ability to model a distribution over plausible scene explanations. Previous optimization-based approaches progressively densify sparse observations with generated novel views and distill them back into 3D~\citep{wu2024reconfusion,sun2025dimensionx,wu2025difix3d+,yu2025viewcrafter}. To avoid their costly per-scene iterative optimization, feed-forward generative methods mainly follow two paradigms: (i) \emph{generation then reconstruction} (\cref{fig:pipeline_comparison}b): it first generates novel views and subsequently uses a feed-forward network to regress a 3D Gaussian splatting (3DGS) representation~\citep{kerbl20233dgs} from them~\citep{szymanowicz2025bolt3d}; (ii) \emph{reconstruction then generation} (\cref{fig:pipeline_comparison}c): it first reconstructs 3DGS and then renders them at target viewpoints to condition a generative model for target-view refinement or synthesis~\citep{chen2024mvsplat360,wewer2024latentsplat,stracquadanio2026reconsplat}.

\input{figures/pipeline_comparison}

However, these feed-forward generative approaches still face several limitations. First, they decouple generation and reconstruction, preventing the two stages from being jointly optimized within a unified 3D representation. For \emph{generation then reconstruction}, inconsistencies across generated views can introduce artifacts into the subsequent 3D reconstruction. For \emph{reconstruction then generation}, the errors in the initial reconstruction are propagated into the generative stage, while the refined target views cannot directly revise the feed-forward 3D reconstruction. Moreover, compared with regression-based methods, these approaches may lose fine-grained geometric and appearance fidelity due to generation in a compressed VAE latent space~\citep{rombach2022sd}, while substantially sacrificing inference efficiency because of their reliance on heavyweight generative models~\citep{rombach2022sd,blattmann2023svd,wan2025wan}.

To address these limitations, we introduce \ours{}, an end-to-end trainable framework that integrates lightweight pixel-space generative modeling directly into feed-forward 3D Gaussian reconstruction. \ours{} consists of a deterministic scene encoder and a generative decoder. Shown in \cref{fig:pipeline_comparison}d, instead of encoding context observations into an explicit 3D reconstruction that may already contain errors, the scene encoder produces a 3D-aware latent, deferring uncertain 3D decisions to generative decoding. Conditioned on this latent and noisy target views, the generative decoder predicts at each sampling step a global 3D Gaussian representation that supports both the observed context views and evolving target views, directly shaping a cross-view-consistent 3D scene. To avoid the representation bottleneck of latent-space generation, \ours{} performs flow matching directly in pixel space~\citep{li2026jit}, where renderings of the predicted 3D Gaussians serve as denoised target predictions, allowing the generative objective to directly supervise 3D reconstruction through differentiable rendering. For efficient generative inference, \ours{} adopts a lightweight DiT~\citep{peebles2023dit} decoder, while the input scene is encoded only once and cached throughout sampling.

Experiments show that this lightweight probabilistic formulation substantially improves reconstruction quality under high uncertainty while retaining sub-second inference. \ours{} outperforms regression-based approaches in challenging settings, including widely spaced input views and view extrapolation, and is even competitive  with methods that rely on much heavier pretrained generative models.
Our main contributions are summarized as follows:
\begin{itemize}
    \item We introduce \ours{}, a highly efficient and end-to-end trainable framework that formulates generalizable feed-forward 3D reconstruction in a probabilistic manner by integrating lightweight generative modeling directly into 3D Gaussian reconstruction.
    \item  We develop pixel-space flow formulation conditioned on a 3D-aware scene latent, avoiding compressed latent-space bottlenecks and enabling direct pixel-space supervision of 3D reconstruction through differentiable rendering.
    \item Extensive experiments show that \ours{} consistently improves over regression-based reconstruction in challenging sparse-view settings, while remaining competitive with much heavier generative approaches and achieving sub-second inference.
\end{itemize}

%% file: figures/pipeline_comparison.tex
\begin{figure*}[t]
    \centering
    \includegraphics[width=\linewidth]{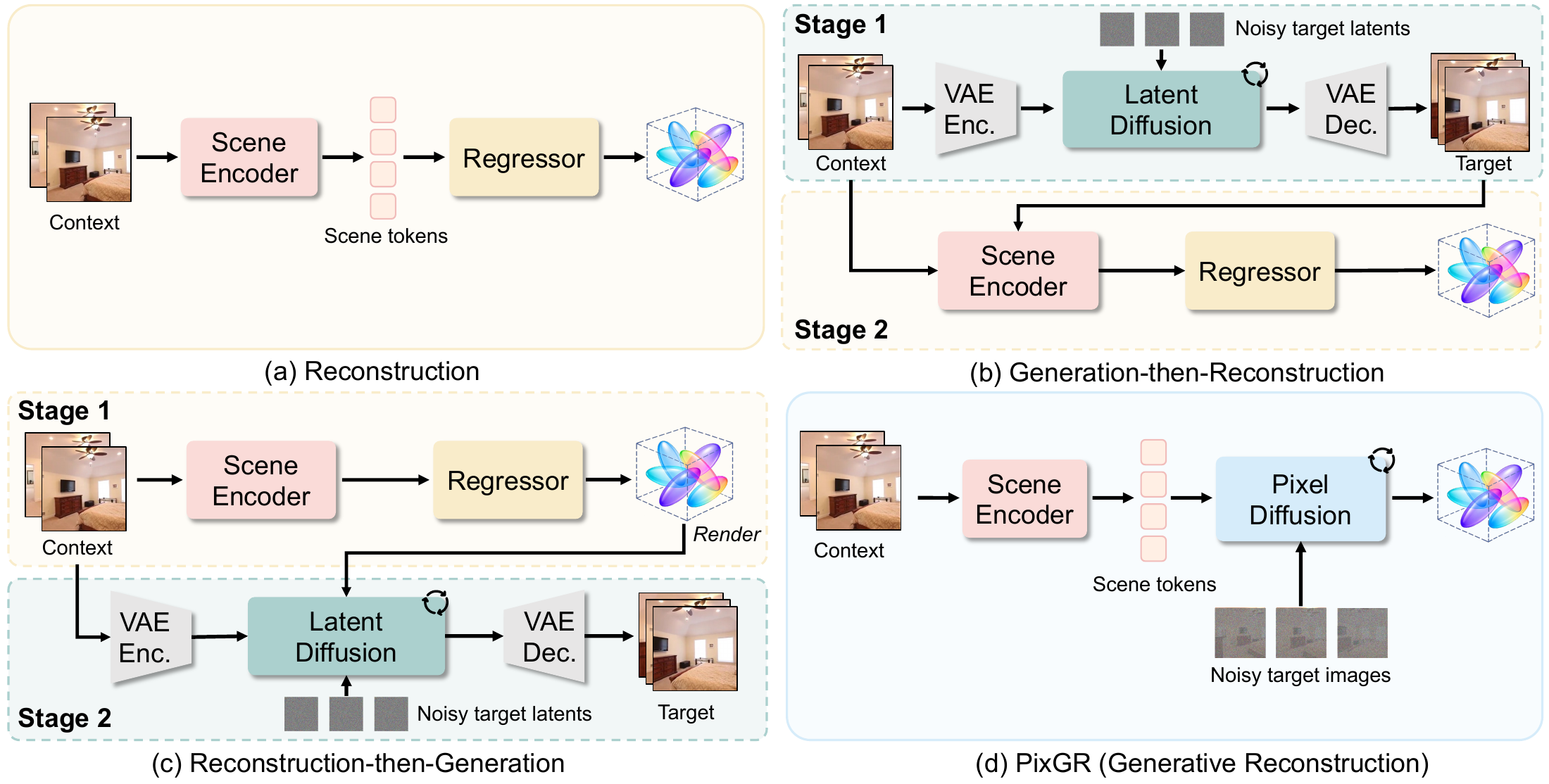}
    \vspace{-18pt}
\caption{Comparison of feed-forward reconstruction and generative pipelines. (a) Deterministic reconstruction cannot model scene uncertainty. Existing generative approaches address this limitation through (b) generation then reconstruction or (c) reconstruction then generation, but suffer from error propagation between stages, VAE-based information loss, and high computational costs. (d) PixG3R instead integrates lightweight pixel-space generation directly into 3D Gaussian reconstruction, enabling end-to-end optimization with direct
pixel-space generative supervision.}
    \label{fig:pipeline_comparison}
    \vspace{-20pt}
\end{figure*}

%% file: secs/related.tex
\section{Related Work}
\label{sec:related-work}

\subsection{Feed-forward 3D Regression}
Recent advancements in feed-forward 3D regression aim to directly regress 3D representations in a single forward pass (\cref{fig:pipeline_comparison}a)~\citep{charatan2024pixelsplat, chen2024mvsplat, wang2024dust3r, huang2025spfsplat, wang2025vggt, xu2025depthsplat, ye2025noposplat, ye2026yonosplat}, significantly improving efficiency compared to per-scene optimization pipelines~\citep{schonberger2016sfm, kerbl20233dgs}. Among these approaches, methods based on 3D Gaussian Splatting (3DGS)~\citep{kerbl20233dgs} are particularly relevant to our work due to their strong rendering efficiency and ability to produce high-quality novel view synthesis.
Within this paradigm, prior work explores a variety of strategies, including incorporating geometric inductive biases~\citep{charatan2024pixelsplat, chen2024mvsplat}, leveraging pretrained depth priors~\citep{xu2025depthsplat}, enabling pose-free inputs~\citep{huang2025spfsplat, huang2025spfsplatv2,ye2025noposplat, ye2026yonosplat}, and adopting self-supervised training schemes~\citep{zhao2026erayzer,huang2026nas3r}.
Despite these advances, feed-forward regression-based methods lack the ability to model uncertainty, often producing blurry reconstructions or failures in occluded and unseen regions.

\subsection{3D Generation}
A line of work learns generative models directly in native 3D representations, including point clouds~\citep{nichol2022pointe}, voxel grids~\citep{muller2023diffrf}, triplanes~\citep{shue20233triplane}, and 3D Gaussians~\citep{zhang2024gaussiancube}, or in structured 3D latent spaces~\citep{lin2025diffsplat,xiang2025trellis}. These methods model 3D distributions directly, but typically rely on large-scale 3D supervision and are mostly developed for bounded object-centric domains.

To exploit 2D generative priors, many methods follow \emph{generation then reconstruction} (\cref{fig:pipeline_comparison}b): they first synthesize novel views using image or video diffusion~\citep{chan2023genvs,liu2023zero123,gao2024cat3d,hollein2024viewdiff,long2024wonder3d,sargent2024zeronvs,shi2024mvdream,wu2024reconfusion,he2025cameractrl,sun2025dimensionx,yu2025viewcrafter,yu2025wonderworld,liu2026reconx}, and then recover 3D through per-scene optimization~\citep{mildenhall2021nerf,kerbl20233dgs} or feed-forward reconstruction~\citep{hong2024lrm,tang2024lgm,xu2024grm,zhang2024gslrm,szymanowicz2025bolt3d}. Although effective, per-scene optimization is computationally expensive, while feed-forward variants remain sensitive to inconsistencies and artifacts across generated views, which can propagate into the reconstructed geometry.
Another paradigm follows \emph{reconstruction then generation} (\cref{fig:pipeline_comparison}c): an initial 3D representation is first reconstructed and then rendered to condition a generative model~\citep{chen2024mvsplat360,wewer2024latentsplat,stracquadanio2026reconsplat}. Such designs generate novel views conditioned on a potentially imperfect 3D reconstruction and cannot directly revise errors in the initial geometry.

Some approaches instead embed differentiable rendering into the generative process~\citep{anciukevivcius2023renderdiffusion,xu2024dmv3d,anciukevivcius2024denoising,cai2025diffusiongs}, but focus mainly on synthetic object-centric scenarios and repeatedly process context observations during generation. In contrast, \ours{} targets real-world scene reconstruction and performs pixel-space flow matching with a lightweight decoder conditioned on a cached 3D-aware scene latent, providing consistent scene guidance across sampling steps while enabling efficient iterative 3D reconstruction.


\subsection{Diffusion and Flow Matching}
Diffusion models~\citep{ho2020ddpm,song2021ddim} and flow matching~\citep{albergo2023buildingflow,lipman2023flow,liu2023flow} have emerged as fundamental frameworks in modern generative modeling, both learning probability flows from a simple initial distribution to a target data distribution. While latent-space methods now dominate high-resolution image generation~\citep{rombach2022stablediffusion,peebles2023dit,ma2024sit}, early diffusion models operated in pixel space with computationally expensive U-Net backbones~\citep{ronneberger2015unet,ho2020ddpm,dhariwal2021diffusion,nichol2021improved,song2021scorebased}. Recent works revisit pixel-space modeling with Vision Transformers (ViT)~\citep{dosovitskiy2021vit,chen2025pixelflow,chen2026dip,li2026jit,yu2026pixeldit,lu2026onestep}; we are particularly inspired by JiT~\citep{li2026jit}, which predicts denoised images that are hypothesized to lie on a low-dimensional manifold~\citep{zhu2005semi}.

%% file: secs/preliminary.tex
\section{Preliminaries}
\label{sec:preliminaries}

\ours{} builds upon the flow-based paradigm~\citep{albergo2023buildingflow, lipman2023flow, liu2023flow} and uses 3D
Gaussians as the scene representation. We therefore briefly review the necessary background on pixel-space flow matching, its multi-view extension, and the 3D Gaussian representation before presenting our full method.

\noindent\textbf{Pixel-Space Flow Matching with $\boldsymbol{x}$-prediction.} 
Consider a data distribution $\vx \sim p_\text{data}(\vx)$, a noise distribution $\ve \sim p_\text{noise}(\ve)$ (\eg, $\ve\sim\mathcal{N}(\boldsymbol{0},\mathbf{I})$), and a timestep $t \in [0, 1]$. During training, a noisy
sample $\vz_t$ is interpolated using a linear schedule as: $
\vz_t = t \, \vx + (1 - t)\, \ve$,
which leads to $\vz_t\sim p_\text{data}$ when $t{=}1$.
A flow velocity $\vv$ is defined as the time-derivative of $\vz$, that is
$\vv = \vx - \ve =(\vx - \vz_t) / (1-t)$.
Following $\vx$-prediction in JiT~\citep{li2026jit}, the network directly predicts the denoised image $\vx_\theta$, with the predicted velocity $\vv_\theta = (\vx_\theta - \vz_t)/(1 - t)$.
Then the velocity loss can be defined as:
\begin{equation}
\mathcal{L} = \mathbb{E}_{t,\vx,\ve} \|  \vv_\theta(\vz_t, t) - \vv \|^2 = \mathbb{E}_{t,\vx,\ve} \frac{1}{(1-t)^2} \| \vx_\theta(\vz_t, t) - \vx \|^2,
\label{eq:loss_v}
\end{equation}
which is equivalent to a time-weighted squared $\ell_2$ image reconstruction loss.
Given the predicted $\vv_\theta$, sampling is done by solving an ordinary differential equation (ODE) $\vz_t$ \citep{albergo2023buildingflow,lipman2023flow, liu2023flow}:
\begin{equation}
{{d\vz_t} / {d t}} = \vv_\theta(\vz_t, t),
\label{eq:ode}
\end{equation}
starting from $\vz_0 \sim p_\text{noise}$ and ending at $t=1$. In practice, it can be approximately solved using numerical solvers such as the Euler and Heun methods.

\noindent\textbf{Multi-View Flow Matching with $\boldsymbol{x}$-prediction.}
The $\boldsymbol{x}$ distribution in pixel-space flow matching is the single image distribution. We extend it to model \emph{joint} distribution of multi-view images $\mathcal{I}=\{\boldsymbol{I}^{i}\}_{i=1}^{V}$, representing $V$ observations of the same 3D scene from viewpoints $\mathcal{P} = \{\boldsymbol{P}^i\}_{i=1}^{V}$. The forward process diffuses every view independently under a shared noise schedule:
\begin{equation}
    \mathcal{I}_{t} = \bigl\{\,t\,\boldsymbol{I} + (1-t)\,\ve_{\boldsymbol{I}} \;\bigm|\; \boldsymbol{I} \in \mathcal{I} \,\bigr\},
\label{eq:diffuse}
\end{equation}
where $\ve_{\boldsymbol{I}} \sim \mathcal{N}(\boldsymbol{0}, \mathbf{I})$ is independent Gaussian noise sampled per view.

\noindent \textbf{3D Gaussian Representation.}
We represent a scene using pixel-aligned 3D Gaussian primitives. For $V$ views of resolution $H \times W$, the Gaussian representation is defined as
$\boldsymbol{G} \in \mathbb{R}^{V \times H \times W \times K}$, where $K$ is the parameter dimension. Each primitive is parameterized by a center $\boldsymbol{\mu} \in \mathbb{R}^{3}$, rotation quaternion $\boldsymbol{r} \in \mathbb{R}^{4}$, scale $\boldsymbol{s} \in \mathbb{R}^{3}$, opacity $\alpha \in \mathbb{R}$, and spherical harmonics coefficients $\boldsymbol{c} \in \mathbb{R}^{k}$.

%% file: secs/methods.tex
\section{Method}
\label{sec:method}
\input{figures/overview}
As shown in \cref{fig:overview}, \ours{} is a fast feed-forward generative reconstruction model that aims to produce a unified 3D Gaussian representation covering both context and target viewpoints. Let $\mathcal{I}_{\sC} = \{\boldsymbol{I}^i_{\sC}\}_{i=1}^{V_{\sC}}$ denote a set of $V_{\sC}$ clean context images with corresponding camera parameters $\mathcal{P}_{\sC} = \{\boldsymbol{P}^i_{\sC}\}_{i=1}^{V_{\sC}}$, and let $\mathcal{P}_{\sT} = \{\boldsymbol{P}^j_{\sT}\}_{j=1}^{V_{\sT}}$ denote the camera parameters of $V_{\sT}$ target views.

\subsection{Reconstruction-based Denoising}
\label{sec:multi-view_flow_matching}
\noindent \textbf{Training.}
Inspired by ~\citep{xu2024dmv3d}, we leverage 3D Gaussian reconstruction and rendering to achieve multi-view image denoising.
During training, we construct noisy target images ${\mathcal{I}}_{\sT}^{(t)}=\{\boldsymbol{I}^{j,(t)}_{\sT}\}_{j=1}^{V_{\sT}}$ by corrupting clean target images $\mathcal{I}_{\sT}=\{\boldsymbol{I}^{j}_{\sT}\}_{j=1}^{V_{\sT}}$ with Gaussian noise at timestep $t$  following \cref{eq:diffuse}.
Our model, denoted as $\mathcal{F}(\cdot)$, outputs pixel-aligned 3D Gaussians $\boldsymbol{G}$ given the clean context images, noisy target images, and their camera parameters:
\begin{equation}
    \boldsymbol{G} = \mathcal{F}\!\left(\mathcal{I}_{\sC},\, \mathcal{P}_{\sC},\, {\mathcal{I}}_{\sT}^{(t)},\, \mathcal{P}_{\sT},\, t\right),
    \label{eq:model}
\end{equation}
Denoising ${\mathcal{I}}_{\sT}^{(t)}$ is performed by rendering $\boldsymbol{G}$ at viewpoints $\mathcal{P}_{\sT}$, yielding the noise-free prediction ${\mathcal{I}}_{\sT}^{(1)}$. This formulation is equivalent to $\vx$-prediction in JiT, allowing the velocity objective in Eq.~\ref{eq:loss_v} to be applied directly to target views. We additionally supervise the model with image reconstruction losses on context and auxiliary novel views to further enforce geometric consistency, and apply perceptual supervision~\citep{zhang2018lpips} to all views. The following loss is applied to all views:

\begin{equation}
    \mathcal{L}
    =
    \mathbb{E}_{(\boldsymbol{I},\boldsymbol{P})}
    \left[
        \omega(\boldsymbol{I},t)
        \left\|
        \mathcal{R}(\boldsymbol{G},\boldsymbol{P})-\boldsymbol{I}
        \right\|^2
        +
          \lambda_{\mathrm{lpips}}
        \operatorname{LPIPS}\!\left(
        \mathcal{R}(\boldsymbol{G},\boldsymbol{P}),\boldsymbol{I}
        \right)\right]\!, 
    \label{eq:loss}
\end{equation}
where $\omega(\boldsymbol{I},t)=(1-t)^{-2}$ for target views $\boldsymbol{I}\in\mathcal{I}_{\sT}$ and $\omega(\boldsymbol{I},t)=1$ otherwise, $\mathcal{R}(\cdot)$ denotes the differentiable renderer, and $\lambda_{\mathrm{lpips}}$ is set to $0.1$.This allows \ours{} to be supervised by 2D image losses in an end-to-end manner. The training process is summarized in \Cref{alg:training}.

\noindent\textbf{Inference.}
At inference, the target images are initialized from pure Gaussian noise $\e \sim \mathcal{N}(\mathbf{0}, \mathbf{I})$, and both the noisy target images and the 3D Gaussians are jointly iterated through the ODE process in \cref{eq:ode}. The final Gaussians are returned as the 3D output.  The full procedure is detailed in \Cref{alg:inference}.

 \subsection{Architecture}
 \label{sec:architecture}
Our model $\mathcal{F}$ consists of a deterministic scene encoder $\mathcal{E}$ and a generative Gaussian decoder $\mathcal{D}$:
\begin{equation}
     \boldsymbol{S} = \mathcal{E}\!\left(\mathcal{I}_{\sC},\, \mathcal{P}_{\sC}\right), \quad
    \boldsymbol{G} = \mathcal{D}\!\left(\boldsymbol{S},\, {\mathcal{I}}_{\sT}^{(t)},\, \mathcal{P}_{\sT},\, t\right),
    \label{eq:model_2}
\end{equation}
where $\boldsymbol{S} \in \mathbb{R}^{V_{\sC} \times P \times d}$ denotes the latent scene representation encoded from context views, with $P$ patch tokens per view and feature dimension $d$.

\noindent\textbf{Scene Encoder.}
\label{sec:encoder}
The scene encoder $\mathcal{E}$ can in principle be instantiated with standard vision transformers~\citep{dosovitskiy2021vit}, as shown in \cref{sec:ablation}. Inspired by~\citep{szymanowicz2026lagernvs}, we establish and initialize $\mathcal{E}$ as the transformer backbone of VGGT~\citep{wang2025vggt}, which is pre-trained for 3D reconstruction under explicit 3D supervision, to provide a 3D-aware scene latent for target-view denoising.

Following VGGT~\citep{wang2025vggt}, all cameras are expressed relative to the first context image $\boldsymbol{I}^1_{\sC}$. The camera parameters for each context view are parameterized as:
    $\boldsymbol{P}^i_{\sC} = \bigl[\mathbf{q}_i,\, \mathbf{t}_i,\, \mathbf{k}_i \bigr],$
where $\mathbf{q}_i \in \mathbb{R}^4$ is the rotation quaternion, $\mathbf{t}_i \in \mathbb{R}^3$ is the camera translation, and $\mathbf{k}_i \in \mathbb{R}^2$ denotes the horizontal and vertical fields-of-view. The camera parameters are projected to 1024-dimensional embeddings using a 2-layer MLP and added to VGGT's learnable camera tokens. We concatenate the patch tokens from the final local and global attention layers along the channel dimension, then project them to dimension $d$ with a linear layer followed by LayerNorm~\citep{ba2016layernorm}. The resulting scene latent $\boldsymbol{S}$ encodes geometry and appearance information from the context observations and serves as the conditioning signal for the generative decoder. Crucially, $\mathcal{E}$  is independent of the flow timestep $t$ and can therefore be computed once and cached throughout the ODE solve, amortizing the cost of the scene encoder across sampling steps.

\noindent\textbf{Generative Gaussian Decoder.}
\label{sec:decoder}
The decoder $\mathcal{D}$ maps observed context views and noisy target views at timestep $t$ to pixel-aligned 3D Gaussians. Following~\citep{xu2024dmv3d,zhang2024gslrm}, each camera $\boldsymbol{P}^{j}$ is encoded as a Pl\"ucker ray map. The corresponding image is concatenated channel-wise with its Pl\"ucker ray map, resulting in a 9-channel posed image that provides per-pixel camera conditioning. 
After patchification and linear projection, each target view is represented as a sequence of patch tokens. The full set of target tokens is stacked as $\boldsymbol{T} \in \mathbb{R}^{V_{\sT} \times P \times d}$.

The scene latent $\boldsymbol{S} $ and target tokens $\boldsymbol{T}$ are concatenated along the view dimension to form a unified token sequence
$\boldsymbol{F} \in \mathbb{R}^{V \times P \times d}$, where $V=V_{\sC} + V_{\sT}$. This sequence is processed by a series of DiT blocks~\cite{peebles2023dit} with AdaLN-Zero conditioning on timestep $t$. Following~\cite{yu2026pixeldit}, we replace LayerNorm with RMSNorm~\cite{zhang2019rmsnorm} and apply 2D RoPE~\cite{su2024roformer} in all attention layers.
Self-attention is performed globally over the entire token sequence $\boldsymbol{F}$, facilitating cross-view information aggregation and enforcing multi-view consistency. 
The refined tokens $\boldsymbol{F}$ are projected to Gaussian parameters via a linear layer and unpatchified to produce pixel-aligned Gaussians $\boldsymbol{G}$. Specifically, we project the tokens to per-view Gaussian parameters and depth maps, with the depth maps lifted to 3D Gaussian centers using the corresponding context and target camera parameters.

Overall, the architecture conditions lightweight global generative decoding with 3D-aware scene features, and is trained end-to-end with \cref{eq:loss} to produce cross-view-consistent 3D Gaussians while denoising target views across sampling steps. 

%% file: figures/overview.tex
\begin{figure*}[t]
  \centering
  \includegraphics[width=\linewidth]{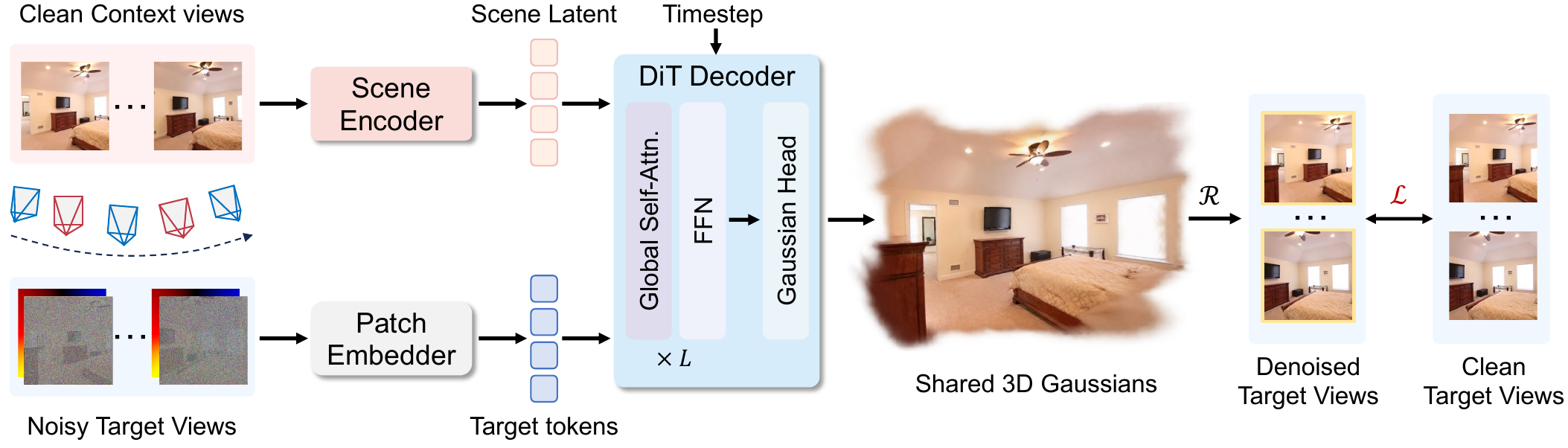}
  \vspace{-10pt}
  \caption{\ours{} Overview.
Multi-view posed context images are encoded once into a scene latent representation. Target viewpoints are sampled and encoded as ray maps, which are combined with noisy target images. A lightweight generative decoder processes the concatenated scene latent and target tokens to predict a shared set of pixel-aligned 3D Gaussians supporting both context and target views. The decoder adopts a DiT-style architecture with global self-attention, followed by a linear Gaussian prediction head. Differentiable rendering of the predicted Gaussians produces denoised target views, enabling direct supervision through pixel-space generative objectives.
  }
  \label{fig:overview}
  \vspace{-5pt}
\end{figure*}

%% file: secs/experiments.tex
\section{Experiments}
\label{sec:experiments}

\subsection{Experimental Setup}
\label{sec:exp_setup}
\noindent\textbf{Implementation Details.}
\ours{} is implemented in PyTorch and trained on A100 GPUs at $256\times 256$ resolution, using AdamW with linear warmup and cosine decay. The logit-normal noise schedule is set to $\mu_t = -0.8$, $\sigma_t = 0.8$, biasing training toward higher noise levels. The generative decoder is 12 DiT blocks with hidden dimension $768$ and is trained from scratch. At inference, we use an Euler sampler. Further details are provided in \cref{app:additional_details}.

\noindent\textbf{Metrics.}
We employ standard novel view synthesis (NVS) metrics: PSNR, SSIM~\citep{wang2004ssim}, LPIPS~\citep{zhang2018lpips}, and FID~\citep{heusel2017fid}, capturing pixel-level accuracy, structural similarity, perceptual quality, and distributional alignment, respectively.

\noindent\textbf{Datasets.}
We train and evaluate on RealEstate10K (RE10K)~\citep{zhou2018stereo} and outdoor DL3DV~\citep{ling2024dl3dv}, and further conduct out-of-domain evaluation on the outdoor ACID dataset~\citep{liu2021infinite} in \cref{app:acid}.

\noindent\textbf{Baselines.}
We compare against two categories of methods. (1) Feed-forward regression-based 3DGS methods, including the pose-required MVSplat~\citep{chen2024mvsplat}, DepthSplat~\citep{xu2025depthsplat}, and YoNoSplat~\citep{ye2026yonosplat}, as well as the pose-free AnySplat~\citep{jiang2025anysplat}. (2) Generative methods, including \emph{generation-then-reconstruction} method  Bolt3D~\citep{szymanowicz2025bolt3d}, and \emph{reconstruction-then-generation} methods MVSplat360~\citep{chen2024mvsplat360} and latentSplat~\citep{wewer2024latentsplat}, ReconSplat~\citep{stracquadanio2026reconsplat}. Additional baseline comparisons are provided in \cref{app:additional_results}, while \cref{app:additional_details} summarizes the geometric and generative priors used by each method and reports their parameter counts.

\noindent\textbf{Target-view selection.}
Unless otherwise specified, \ours{} takes three target keyframe viewpoints as queries and generates a shared 3D Gaussian representation conditioned on the scene latent. We render the resulting representation at all evaluation viewpoints for metric computation.
\subsection{Results}

\input{tables/interpolation}
\input{figures/interpolation}

\subsubsection{View Interpolation}
\label{sec:interpolation}
\cref{tab:interpolation} shows view interpolation on the 2-view RE10K and 6-view DL3DV benchmarks, where the novel cameras lie between the context viewpoints. For RE10K, we use the three evaluation viewpoints directly as target keyframes. For DL3DV, we uniformly sample three target keyframes within the range spanned by the context views, and render the resulting 3D representation from all 42 evaluation viewpoints. As shown in \cref{tab:interpolation}, \ours{} achieves the best performance on most metrics across both datasets. The qualitative comparisons in \cref{fig:interpolation} further show that \ours{} preserves stable geometry, consistent appearance, and fine details across substantial viewpoint changes, while other methods exhibit blurring in challenging regions or strong generation artifacts.

\subsubsection{View Extrapolation}
\label{sec:extrapolation}
We further evaluate 2-view RE10K and 4-view DL3DV on view extrapolation, where novel cameras move beyond the observed range and the model must infer geometry and appearance that are weakly or not at all constrained by the context views. For RE10K, we use its three evaluation poses as targets. For DL3DV, we sample three target poses evenly across the extrapolation range and render the resulting 3DGS at all 42 evaluation poses. As shown in \cref{tab:extrapolation}, \ours{} achieves large gains over regression-based methods across all metrics, highlighting the difficulty of deterministic regression in weakly observed regions. This is also reflected qualitatively in \cref{fig:extrapolation}, where regression baselines exhibit blurring and severe artifacts in unseen or under-constrained areas.

Although previous generative methods improve FID over regression baselines on RE10K, they still struggle to synthesize high-quality extrapolated views. The decoupling between generation and reconstruction can introduce geometry errors and cross-view misalignment, while compressed latent-space generation may lose fine details and cause hue shifts. As shown in \cref{fig:extrapolation}, these methods exhibit geometric errors and detail loss. In contrast, \ours{} performs pixel-space generation directly within the 3D reconstruction process, producing sharper, coherent novel views with finer details.


\input{tables/extrapolation}
\input{figures/extrapolation}

\subsubsection{Efficiency Analysis}
\label{sec:efficiency}
\input{figures/novel_view_time_scaling}
\Cref{fig:novel_view_time_scaling} compares inference time as the number of evaluation novel views increases on a single A100 GPU. We measure the total time to obtain a 3D Gaussian representation and render or generate the requested novel views. Regression-based methods remain fast but are less effective under ambiguous extrapolation (\cref{sec:extrapolation}), while existing generative methods incur substantially higher cost that grows with the number of requested views. In contrast, \ours{} remains below one second from 3 to 48 novel views, benefiting from a lightweight generative decoder and a cached scene representation. This favorable scaling is further enabled by generating a single global 3D Gaussian representation conditioned on the context views and three key target frames, from which all evaluation views can be efficiently rendered without repeated generation. These results demonstrate that our paradigm can model scene uncertainty while preserving the efficiency of feed-forward reconstruction.

\subsubsection{Ablation Study}
\label{sec:ablation}
\input{tables/ablation}
\input{figures/ablation}
\paragraph{Encoder Design.} As shown in \cref{tab:ablation}, removing VGGT initialization or replacing the VGGT-based encoder with a randomly initialized ViT-L degrades all metrics, indicating that the generative decoder benefits from a 3D-aware scene latent. Nevertheless, both variants still outperform most baselines in \cref{tab:interpolation,tab:extrapolation}, demonstrating that the decoder remains effective across different encoder choices and that our paradigm does not rely entirely on pretrained 3D priors.

\paragraph{Generative vs. Deterministic Decoding.} In \cref{tab:ablation}, replacing the DiT decoder with a 12-block deterministic regressor of comparable model size leads to a consistent performance drop under interpolation and a substantially larger drop under extrapolation, where scene uncertainty is higher. This widening gap under extrapolation supports our motivation that deterministic regression is limited when geometry and appearance are weakly constrained by the observations. \Cref{fig:ablation} further demonstrates that the generative decoder recovers sharper appearance and more coherent geometry when context views have limited or no overlap.

\paragraph{Keyframe Selection.} We further study the number of target keyframes used to construct the shared 3D Gaussian representation. As shown in \cref{fig:dl3dv_keyframes}, using more keyframes consistently improves performance, with larger gains under extrapolation where scene ambiguity is higher. Jointly denoising target views at additional camera poses encourages the shared representation to cover a broader viewing range and strengthens cross-view consistency in weakly observed regions. Most of the improvement is obtained with two keyframes, while a third provides a further quality gain at an additional latency of only $0.09$ s. We therefore use three target keyframes by default.

%% file: tables/interpolation.tex
\begin{table*}[t]
    \footnotesize
    \centering
    \caption{Quantitative comparisons of 3D reconstruction performance on view interpolation. Results are reported on 2-view RE10K and 6-view DL3DV. \ours{} consistently outperforms prior methods across datasets. \best{Best} and \secondbest{second-best} results are highlighted.}
    \setlength{\tabcolsep}{11.5pt}
    \renewcommand{\arraystretch}{1.1}
    \begin{tabular}{
        l
        w{c}{0.6cm} w{c}{0.6cm} w{c}{0.6cm} w{c}{0.6cm}
        w{c}{0.6cm} w{c}{0.6cm} w{c}{0.6cm} w{c}{0.6cm}
    }
    \toprule
    \multirow{2}{*}{\raisebox{-0.55ex}{\textbf{Method}}}
    & \multicolumn{4}{c}{\textbf{RE10K}}
    & \multicolumn{4}{c}{\textbf{DL3DV}} \\
    \cmidrule(r{2pt}){2-5}
    \cmidrule(r{2pt}){6-9}
    & PSNR $\uparrow$ 
    & SSIM $\uparrow$
    & LPIPS $\downarrow$
    & FID $\downarrow$
    & PSNR $\uparrow$
    & SSIM $\uparrow$
    & LPIPS $\downarrow$
    & FID $\downarrow$
    \\
    \midrule
    \rowcolor{gray!20}
    \multicolumn{9}{l}{\textit{Regression-based Methods}} \\
    MVSplat
    & 26.39 & 0.869 & 0.128 & 2.28 & 18.12 & 0.556 & 0.370 & 58.20 \\
    DepthSplat
    & \secondbest{27.47} & \secondbest{0.889} & \secondbest{0.114} & 2.29 & 24.19 & \secondbest{0.823} & \secondbest{0.147} &  \best{9.51} \\
    AnySplat 
    & 19.80 & 0.682 & 0.274 & 2.88 & 19.03 & 0.554 & 0.235 & 16.94 \\
    YoNoSplat 
    & 25.73 & 0.851 & 0.130 & 3.75 & \secondbest{24.83} & 0.821 & \best{0.137} & 11.56 \\
    \rowcolor{gray!20}
    \multicolumn{9}{l}{\textit{Generative Methods}} \\
    latentSplat 
    & 23.89 & 0.813 & 0.163 & \secondbest{2.22} & 17.22 & 0.514 & 0.385 & 45.66 \\
    MVSplat360 & 26.41 & 0.869 & 0.126 & -- & 20.07 & 0.698 & 0.257 & 18.68 \\
    ReconSplat & 25.57	& 0.832	& 0.150 &4.29 & 23.05 &	0.728 &	0.205	& 18.93 \\
     \cellcolor{ourscell}\textbf{\ours{}}
    & \cellcolor{ourscell}\best{28.54}
    & \cellcolor{ourscell}\best{0.905}
    & \cellcolor{ourscell}\best{0.104}
    & \cellcolor{ourscell}\best{2.19}
    & \cellcolor{ourscell}\textbf{25.27}
    & \cellcolor{ourscell}\textbf{0.840}
    & \cellcolor{ourscell}\secondbest{0.147}
    & \cellcolor{ourscell}\secondbest{11.26}\\
    \bottomrule
    \end{tabular}
    \label{tab:interpolation}
    \vspace{-5pt}
\end{table*}

%% file: figures/interpolation.tex
\begin{figure*}[t]
  \centering
  \includegraphics[width=\linewidth]{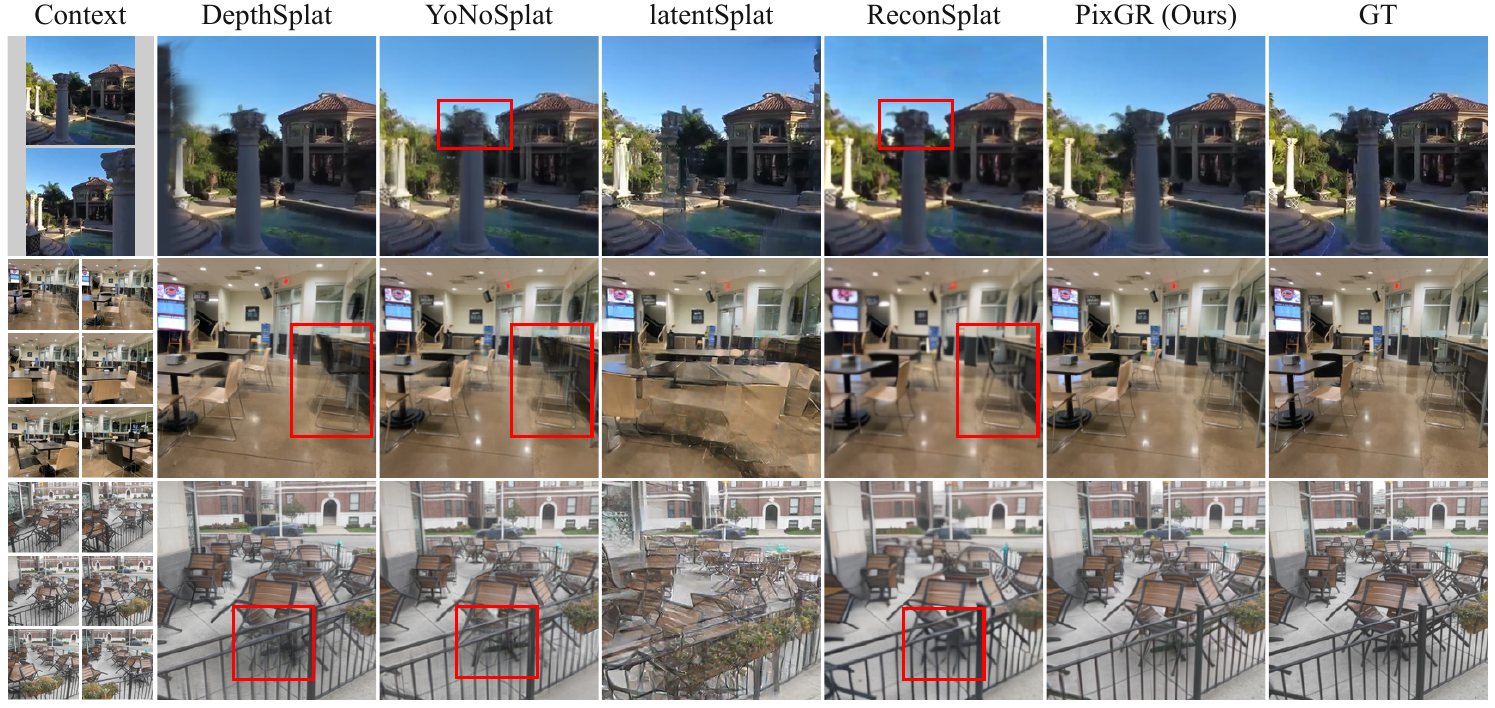}
    \vspace{-20pt}
  \caption{Qualitative comparison on interpolated views. \ours{} produces sharper and more geometrically consistent renderings than prior methods and handles large viewpoint changes effectively.}
  \label{fig:interpolation}
  \vspace{-13pt}
\end{figure*}

%% file: tables/extrapolation.tex
\begin{table*}[t]
    \footnotesize
    \centering
    \caption{Quantitative comparisons of 3D reconstruction performance on view extrapolation. Results are reported on 2-view RE10K and 4-view DL3DV. \ours{} consistently outperforms prior methods.}
    \vspace{-5pt}
    \setlength{\tabcolsep}{12pt}
    \renewcommand{\arraystretch}{1.1}
    \begin{tabular}{
        l
        w{c}{0.6cm} w{c}{0.6cm} w{c}{0.6cm} w{c}{0.6cm}
        w{c}{0.6cm} w{c}{0.6cm} w{c}{0.6cm} w{c}{0.6cm}
    }
    \toprule
    \multirow{2}{*}{\raisebox{-0.55ex}{\textbf{Method}}}
    & \multicolumn{4}{c}{\textbf{RE10K}}
    & \multicolumn{4}{c}{\textbf{DL3DV}} \\
    \cmidrule(r{2pt}){2-5}
    \cmidrule(r{2pt}){6-9}
    & PSNR $\uparrow$
    & SSIM $\uparrow$
    & LPIPS $\downarrow$
    & FID $\downarrow$
    & PSNR $\uparrow$
    & SSIM $\uparrow$
    & LPIPS $\downarrow$
    & FID $\downarrow$ \\
    \midrule
    \rowcolor{gray!20}
    \multicolumn{9}{l}{\textit{Regression-based Methods}} \\
    DepthSplat
    & 22.02 & \secondbest{0.822} & 0.189 & 6.95
    & 21.59 & 0.745 & 0.205 & 17.41 \\
    AnySplat
    & 16.21 & 0.621 & 0.351 & 5.98
    & 14.84 & 0.449 & 0.426 & 22.16 \\
    YoNoSplat
    & 23.00 & 0.810 & 0.182 & 6.47
    & \secondbest{23.04} & \secondbest{0.761} & \secondbest{0.192} & \secondbest{17.21} \\
    \rowcolor{gray!20}
    \multicolumn{9}{l}{\textit{Generative Methods}} \\
    latentSplat
    & 22.62 & 0.777 & 0.196 & {2.79}
    & 18.35 & 0.564 & 0.345 & 37.65 \\
    MVSplat360
    & \secondbest{23.16} & 0.810 & 0.176 & \best{1.79}
    & 17.91 & 0.609 & 0.335 & 18.23 \\
    Bolt3D
    & 23.13 & 0.806 & \secondbest{0.166} & 3.66
    & -- & -- & -- & -- \\
    ReconSplat & 23.98	& 0.795	& 0.184 &4.49 &21.23	& 0.673 & 0.249	& 23.02 \\
    \cellcolor{ourscell}\textbf{\ours{}}
    & \cellcolor{ourscell}\best{26.03}
    & \cellcolor{ourscell}\best{0.872}
    & \cellcolor{ourscell}\best{0.141}
    & \cellcolor{ourscell}\secondbest{2.75}
    & \cellcolor{ourscell}\best{23.76}
    & \cellcolor{ourscell}\best{0.802}
    & \cellcolor{ourscell}\best{0.178}
    & \cellcolor{ourscell}\best{12.78} \\
    \bottomrule
    \end{tabular}
    \label{tab:extrapolation}
    \vspace{-10pt}
\end{table*}

%% file: figures/extrapolation.tex
\begin{figure*}[t]
  \centering
  \includegraphics[width=\linewidth]{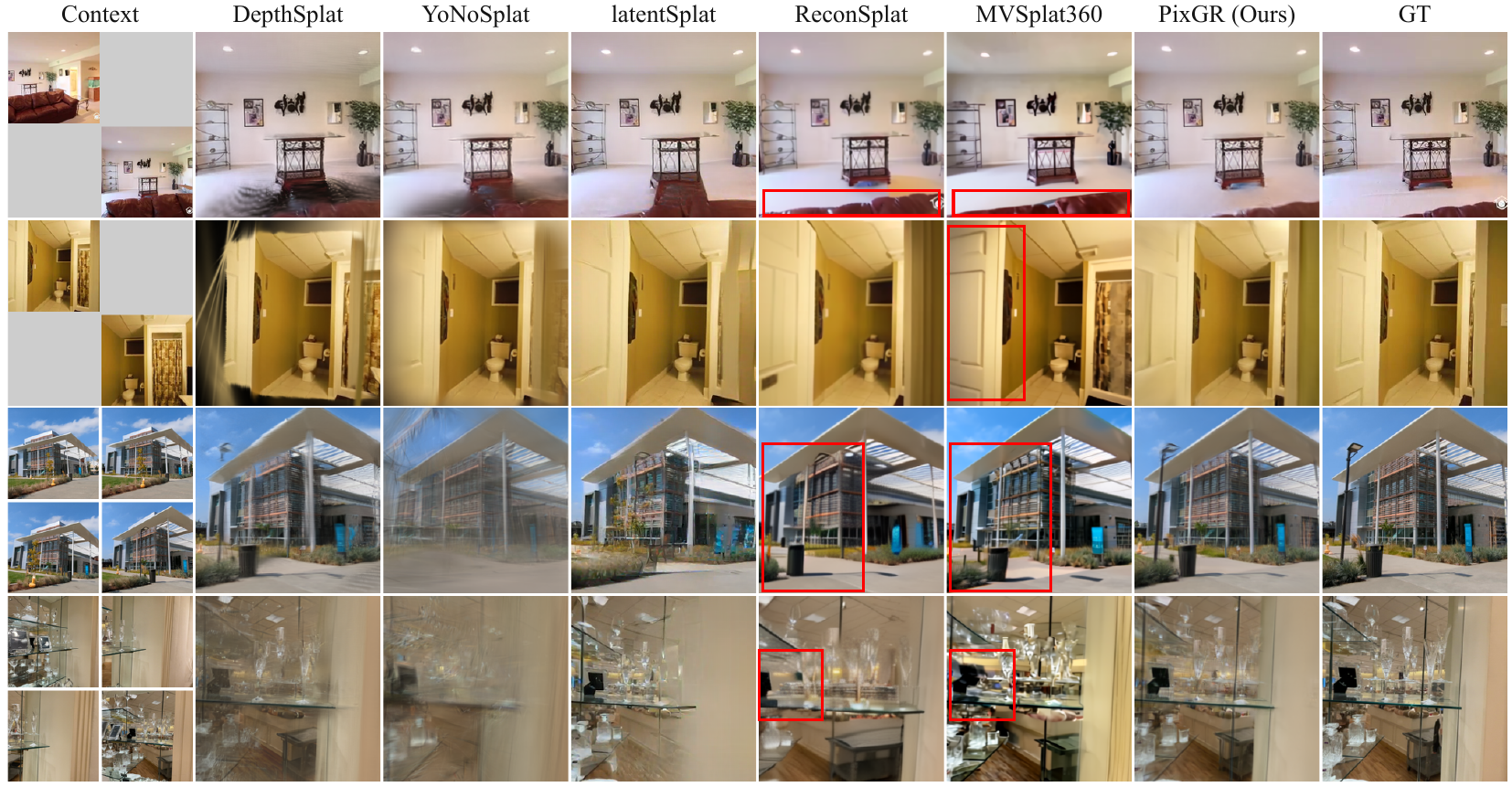}
    \vspace{-15pt}
  \caption{Qualitative comparison on extrapolated novel views. Regression-based methods fail in unseen regions, while generative baselines may produce artifacts and distortions, and lead to geometric misalignment (e.g., the sofa in Row 1) and detail loss (e.g., missing texture on the dustbin and the missing street lamp in Row 3). In contrast, \ours{} produces geometrically consistent renderings while preserving fine details.}
  \label{fig:extrapolation}
  \vspace{-15pt}
\end{figure*}

%% file: figures/novel_view_time_scaling.tex
\begin{wrapfigure}{r}{0.43\columnwidth}
    \centering
    \vspace{-\intextsep}
    \includegraphics[width=\linewidth]{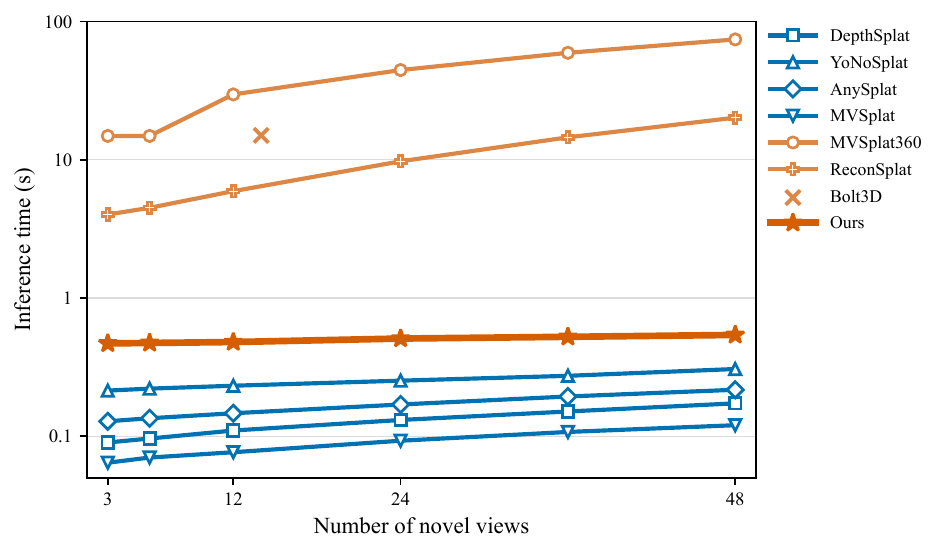}
     \vspace{-15pt}
    \caption{Inference time versus the number of novel views on a single A100 GPU. }
      \vspace{-10pt}
    \label{fig:novel_view_time_scaling}
\end{wrapfigure}

%% file: tables/ablation.tex
\begin{table*}[t]
    \footnotesize
    \centering
    \caption{Ablation of encoder and decoder design on RE10K under interpolation and extrapolation. PixGR uses a pretrained VGGT encoder and generative decoder, achieving the best performance.}
    \vspace{-5pt}
    \setlength{\tabcolsep}{9.5pt}
    \renewcommand{\arraystretch}{1.1}
    \begin{tabular}{
        l
        w{c}{0.7cm} w{c}{0.7cm} w{c}{0.7cm} w{c}{0.7cm}
        w{c}{0.7cm} w{c}{0.7cm} w{c}{0.7cm} w{c}{0.7cm}
    }
    \toprule
    \multirow{2}{*}{\raisebox{-0.55ex}{\textbf{Variant}}}
    & \multicolumn{4}{c}{\textbf{Interpolation}}
    & \multicolumn{4}{c}{\textbf{Extrapolation}} \\
    \cmidrule(r{2pt}){2-5}
    \cmidrule(r{2pt}){6-9}
    & PSNR $\uparrow$
    & SSIM $\uparrow$
    & LPIPS $\downarrow$
    & FID $\downarrow$
    & PSNR $\uparrow$
    & SSIM $\uparrow$
    & LPIPS $\downarrow$
    & FID $\downarrow$ \\
    \midrule
    \rowcolor{gray!20}
    \multicolumn{9}{l}{\textit{Encoder ablation}} \\
    VGGT w/o init.
    & 27.32 & 0.882 & 0.134 & 2.45
    & 24.98 & 0.859 & 0.167 & 3.67 \\
    ViT-L encoder
    & 27.14	& 0.881 & 0.126 & 2.63
    & 24.80 & 0.848 & 0.170 & 3.89 \\
    \rowcolor{gray!20}
    \multicolumn{9}{l}{\textit{Decoder ablation}} \\
    Regressor
    & 27.58 & 0.883 & 0.111 & 2.38
    & 24.34 & 0.821 & 0.166 & 5.42 \\
    \midrule
    \cellcolor{ourscell}\textbf{\ours{}}
    & \cellcolor{ourscell}\best{28.54}
    & \cellcolor{ourscell}\best{0.905}
    & \cellcolor{ourscell}\best{0.104}
    & \cellcolor{ourscell}\best{2.19}
    & \cellcolor{ourscell}\best{26.03}
    & \cellcolor{ourscell}\best{0.872}
    & \cellcolor{ourscell}\best{0.141}
    & \cellcolor{ourscell}\best{2.75} \\
    \bottomrule
    \end{tabular}
    \label{tab:ablation}
    \vspace{-10pt}
\end{table*}

%% file: figures/ablation.tex
\begin{figure}[t]
    \centering
    \begin{minipage}{0.45\textwidth}
        \centering
        \includegraphics[width=\linewidth]{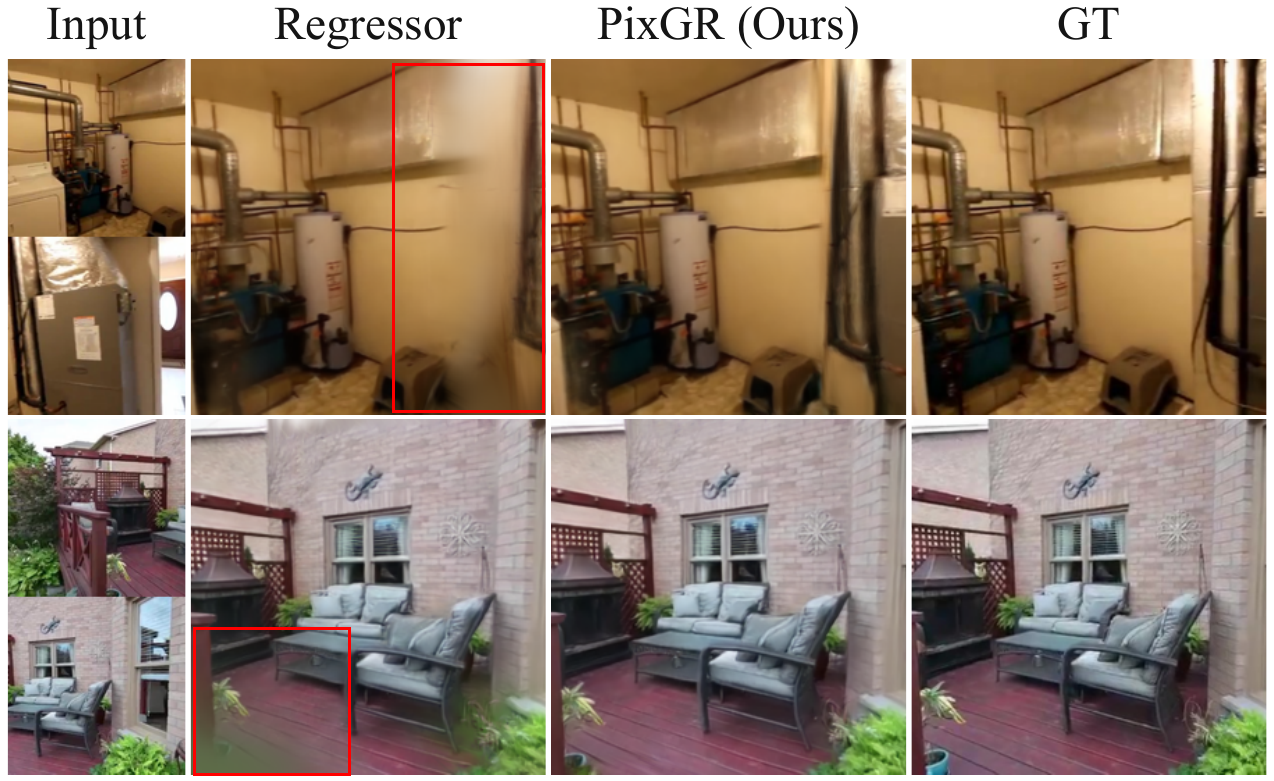}
         \vspace{-20pt}
       \caption{Qualitative comparison between a regressor and DiT as the Gaussian decoder.}
        \label{fig:ablation}
    \end{minipage}
    \hfill
    \input{figures/dl3dv_keyframes}
    \vspace{-10pt}
\end{figure}

%% file: figures/dl3dv_keyframes.tex
\begin{minipage}{0.51\textwidth}
    \centering
    \vspace{7pt}
    \includegraphics[width=\linewidth]{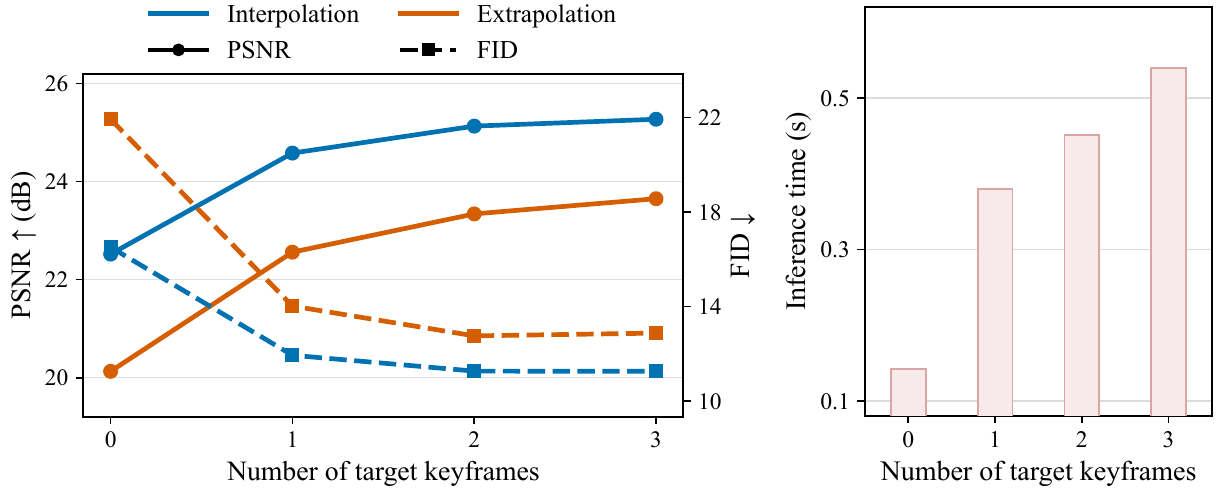}
    \vspace{-10pt}
    \caption{Effect of target keyframes on DL3DV. PSNR and FID (left) and inference time (right) versus the number of target keyframes.}
    \label{fig:dl3dv_keyframes}
\end{minipage}

%% file: secs/conclusion.tex
\section{Conclusion}
\label{sec:conclusion}

We present \ours{}, a novel framework for generative 3D reconstruction that combines a 3D-aware scene encoder with a lightweight generative decoder. Despite using a lightweight decoder trained without pretrained generative priors, \ours{} achieves strong performance in both interpolation and extrapolation, outperforming regression-based methods and remaining competitive with heavier generative approaches. These results show that lightweight pixel-space generation provides an effective and scalable approach to modeling uncertainty in sparse-view 3D reconstruction.

%% file: secs/appendix.tex
\section{Method}
\label{app:method_details}

\Cref{alg:training,alg:inference} summarize the training and inference procedures, respectively.
\input{algorithms/training}

\input{algorithms/inference}

\section{More Implementation Details}
\label{app:additional_details}
\paragraph{Implementation Details.}
\ours{} is trained for $100{,}000$ steps and an initial learning rate of $1\mathrm{e}{-4}$. During training, target views are sampled from an extrapolated range of the context views, so that the target views may lie outside or inside the context range. All cameras in one scene are expressed relative to the first context view, and the camera poses are normalized to a fixed scale across all scenes. 
For interpolation, we adopt the evaluation splits from~\cite{charatan2024pixelsplat,xu2025depthsplat} for 2-view RE10K and ACID (\cref{tab:acid}), and the 6-view DL3DV benchmark. For extrapolation, we adopt the extrapolated evaluation splits from latentSplat for RE10K, and remove the first and last two views from the 6-view DL3DV benchmark to form a 4-view extrapolation setting. We use three target keyframes for both interpolation and extrapolation and 10 sampling steps unless otherwise specified. We sample 4 novel views for auxiliary supervision.

\paragraph{Architecture.}
VGGT operates on images with patch size 14, while our decoder tokenizer processes 256$\times$256 images with patch size 8. To align the token counts between encoder and decoder, we resize the conditioning images to 448$\times$ 448, so that VGGT outputs 1024 (32$\times$ 32) tokens per view, matching the number of target tokens. The per-view tokens are then concatenated and passed to the decoder to produce per-view pixel-aligned Gaussians.
We adopt a lightweight DiT architecture with a hidden dimension of 768, 12 attention heads, and 12 transformer blocks, comprising only 131M parameters, as reported in \cref{tab:regression_priors}.

\paragraph{Implementation Details of Baselines.}
For YoNoSplat, we use its GT-pose- and GT-intrinsics-required setting and evaluate at $256\times 256$. 
As Bolt3D is not open-sourced, we compare to it only quantitatively, reporting the numbers from their paper under the same experimental settings. 

\paragraph{Implementation Details of Efficiency Analysis}
For a fair comparison, all methods use two context views at a resolution of $256\times256$, except AnySplat, for which we use $252\times252$ inputs to accommodate its patch size; this minor difference has a negligible effect on computational cost. Since Bolt3D is not publicly available, we report its published runtime for a fixed input of 16 views as a reference.

\paragraph{Priors and Parameter Counts.}
\input{tables/regression_priors}
\Cref{tab:regression_priors} summarizes the geometric and pretrained generative priors used by the compared methods together with their encoder, decoder, and total parameter counts when available. Here we also include some additional methods such as epth Anything V3 (DA3)~\citep{lin2025depthanythingv3}, NoPoSplat~\citep{ye2025noposplat}, SPFSplat~\citep{huang2025spfsplat} and NAS3R~\citep{huang2026nas3r} which will be evaluated in \cref{tab:additional_regression_extrapolation}. 

Regression-based approaches leverage geometric foundation models but do not incorporate generative priors for resolving ambiguity in extrapolated regions. Existing generative approaches instead build on large pretrained image or video generators, whereas \ours{} combines a VGGT geometric prior with a lightweight pixel-space generative decoder trained from scratch. We also include the \ours{} variant introduced in \cref{sec:ablation}, which replaces the VGGT encoder with ViT-L.


\section{Additional Quantitative Results}
\label{app:additional_results}
\input{tables/acid}
\paragraph{Out-of-Domain Evaluation on ACID.}
\label{app:acid}
As shown in \cref{tab:acid}, \ours{} consistently outperforms both regression-based and generative baselines on the out-of-domain ACID benchmark, achieving the best results across all four metrics. In particular, it surpasses the strongest competing method, MVSplat, by 1.13 dB in PSNR while also improving SSIM, LPIPS, and FID. These results demonstrate the robustness of \ours{} and highlight the effectiveness of pixel-space generative decoding for producing high-fidelity and structurally consistent novel views under previously unseen visual distributions.

\paragraph{Comparison with Additional Regression-Based Methods.}
We mainly compare with pose-require methods in our main experiments, as they are the most relevant to our setting.
We additionally compare with recent pose-free regression-based reconstruction methods, Depth Anything V3 (DA3)~\citep{lin2025depthanythingv3}, NoPoSplat~\citep{ye2025noposplat}, SPFSplat~\citep{huang2025spfsplat} and NAS3R~\citep{huang2026nas3r} on the 2-view RE10K extrapolation setting. We adopt NAS3R's variant that initializes from VGGT. As shown in \cref{tab:additional_regression_extrapolation}, \ours{} substantially outperforms all additional baselines across all metrics, further demonstrating the benefit of generative decoding when novel views lie outside the observed camera range.
\input{tables/additional_regression_extrapolation}

\begin{figure*}[t!]
    \centering
    \begin{minipage}[t]{0.48\textwidth}
        \centering
        \scriptsize
        \setlength{\tabcolsep}{5pt}
        \renewcommand{\arraystretch}{1.0}
        \vspace{-105pt}
        \captionof{table}{One-view reconstruction results on RE10K compared with DiffusionGS~\citep{cai2025diffusiongs}. \best{Best} results are highlighted. Our method consistently outperforms DiffusionGS across all metrics, demonstrating better performance in reconstructing scenes from a single view.}
        \begin{tabular}{lcccc}
            \toprule
            \textbf{Method} & PSNR $\uparrow$ & SSIM $\uparrow$ & LPIPS $\downarrow$ & FID $\downarrow$ \\
            \midrule
            DiffusionGS & 17.09 & 0.560 & 0.420 & 7.88 \\
            \cellcolor{ourscell}\textbf{\ours{}}
            & \cellcolor{ourscell}\best{17.65}
            & \cellcolor{ourscell}\best{0.600}
            & \cellcolor{ourscell}\best{0.384}
            & \cellcolor{ourscell}\best{6.15} \\
            \bottomrule
        \end{tabular}
        \label{tab:one_view_reconstruction}
    \end{minipage}
    \hfill
    \begin{minipage}[t]{0.48\textwidth}
        \centering
        \includegraphics[width=0.9\linewidth]{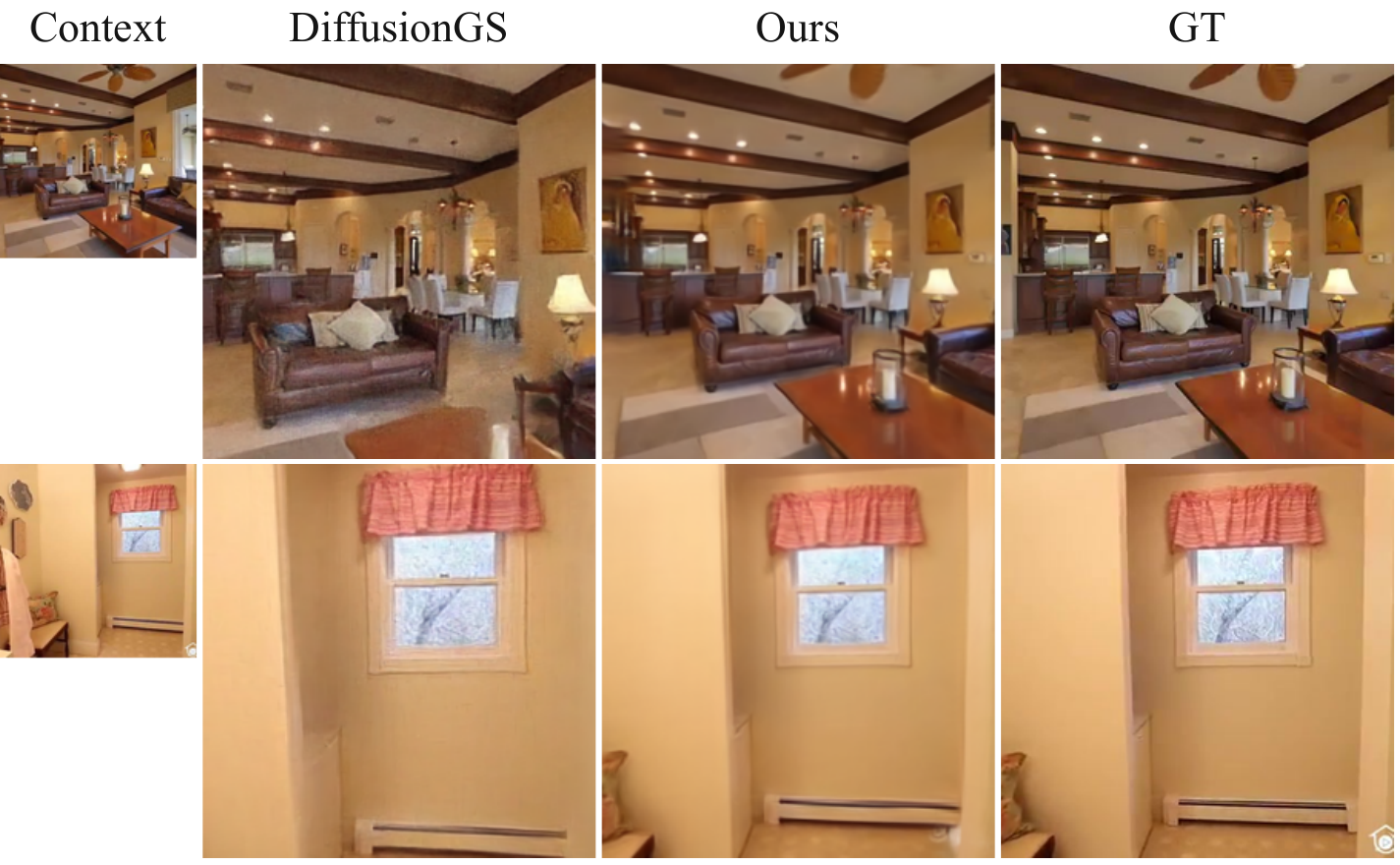}
        \vspace{-10pt}
        \caption{Visualization of one-view reconstruction on RE10K.}
        \label{fig:one_view_reconstruction}
    \end{minipage}
\end{figure*}

\paragraph{One-View Reconstruction on RE10K.}
We further evaluate reconstruction from a single context view, where scene geometry is highly under-constrained. As shown in \cref{tab:one_view_reconstruction}, \ours{} consistently outperforms DiffusionGS~\citep{cai2025diffusiongs}, improving PSNR by $0.56$ dB and reducing FID by $1.73$. The qualitative comparison in \cref{fig:one_view_reconstruction} further shows sharper details and more accurate scene structure, while DiffusionGS often produces renderings that deviate from the target viewpoint. These results demonstrate that conditioning pixel-space generative decoding on a 3D-aware scene representation remains effective even in this extremely sparse setting, providing consistent geometric guidance across sampling steps.


\FloatBarrier

\section{Limitations and Future Work}
\label{app:limitations}
While \ours{} achieves strong performance in both interpolation and extrapolation, it has several limitations. First, \ours{} employs a lightweight generative decoder trained from scratch, which is less expressive than large pretrained generative models. This limits its ability to synthesize high-fidelity novel views in highly ambiguous scenarios, such as long-sequence generation from a single view. Future work could explore incorporating pretrained generative priors while preserving efficient inference. Second, the current 3D Gaussian representation is limited to static scenes and cannot model dynamic changes such as motion or deformation. Extending this paradigm to dynamic scenes with 4D representations is a promising direction for future work.

%% file: algorithms/training.tex
\begin{algorithm}[H]
  \small
  \begin{algorithmic}[1]
  \Require Training set $\mathcal{D}$; Scene encoder $\mathcal{E}$; Generative Gaussian decoder $\mathcal{D}$; Renderer $\mathcal{R}$; number of training iterations $K$.
  \For {$i=1,\ldots,K$}
      \State Draw a batch from the training set: $(\mathcal{I}_{\sC},\mathcal{I}_{\sT},\mathcal{I}_{\sN},\mathcal{P}_{\sC},\mathcal{P}_{\sT},\mathcal{P}_{\sN})
      \sim \mathcal{D}.$
      \State Sample a timestep $t = \sigma\!\left(\xi\right)$, $ \xi \sim \mathcal{N}(\mu_t,\, \sigma_t^2)$ and Gaussian noise $\e \sim \mathcal{N}(0,\mathbf{I})$.
      \State Corrupt clean target images:
      $
      \mathcal{I}_{\sT}^{(t)} = \{\boldsymbol{I}^{(t)}_{\sT} =
      t\boldsymbol{I}_{\sT} + (1-t)\boldsymbol{\epsilon}_v
      \; |\;
      \boldsymbol{I}_{\sT} \in \mathcal{I}_{\sT}\}.
      $
      \State Predict pixel-aligned 3D Gaussians:
      $
      \boldsymbol{S}
      \gets
      \mathcal{E}(\mathcal{I}_{\sC}, \mathcal{P}_{\sC}), \; \boldsymbol{G}
      \gets
      \mathcal{D}
      \left(
      \boldsymbol{S},\, {\mathcal{I}}_{\sT}^{(t)},\, \mathcal{P}_{\sT},\, t\right).
      $
     \State Render images at all viewpoints:
     $    
      \widehat{\mathcal{I}}
      =
      \{
      \hat{\boldsymbol{I}}
      =
      \mathcal{R}(\boldsymbol{G}, \boldsymbol{P})
      \;|\;
      \boldsymbol{P} \in \mathcal{P}_{\sC} \cup \mathcal{P}_{\sT} \cup \mathcal{P}_{\sN}
      \}.
      $
      \State Compute main objective $ \mathcal{L}
    =
    \mathbb{E}_{(\boldsymbol{I},\widehat{\boldsymbol{I}})}
    \left[
        \omega(\boldsymbol{I},t)
        \left\|
        \widehat{\boldsymbol{I}}-\boldsymbol{I}
        \right\|^2
        +
        \lambda_{\mathrm{lpips}}
        \operatorname{LPIPS}\!\left(
        \widehat{\boldsymbol{I}},\boldsymbol{I}
        \right)
    \right]\!,$ where $\omega(\boldsymbol{I},t)=(1-t)^{-2}$ for  $\boldsymbol{I}\in\mathcal{I}_{\sT}$, and $\omega(\boldsymbol{I},t)=1$ for  $\boldsymbol{I}\in\mathcal{I}_{\sC}\cup\mathcal{I}_{\sN}$, and update $\mathcal{E}$, $\mathcal{D}$ via backpropagation.
  \EndFor
  \State \Return $e, h$.
  \end{algorithmic}
  \caption{Training.}
  \label{alg:training}
  \end{algorithm}

%% file: algorithms/inference.tex

\begin{algorithm}[H]
\small
\begin{algorithmic}[1]
    \Require Context images $\mathcal{I}_{\sC}$, Context camera parameters $\mathcal{P}_{\sC}$, Target camera parameters $\mathcal{P}_{\sT}$, Scene encoder $\mathcal{E}$; Generative Gaussian decoder $\mathcal{D}$; Renderer $\mathcal{R}$; number of  denoising steps $N$.
    \State Initialize target images: $
    \mathcal{I}_{\sT}^{(0)}
    =
    \{
    \boldsymbol{I}_{\sT}^{j,(0)}
    \;|\;
    \boldsymbol{I}_{\sT}^{j,(0)}
    \sim
    \mathcal{N}(\mathbf{0}, \mathbf{I})
    \}_{j=1}^{V_\sT}.
    $
    \State Encode scene representation: $\boldsymbol{S} \gets \mathcal{E}\!\left(\mathcal{I}_{\sC},\, \mathcal{P}_{\sC}\right)$.
    \For {$i=0,\ldots,N-1$}
    \State Set timestep: $t_i = i/N$,\, $t_{i+1} = (i{+}1)/N$.
    \State Decode 3D Gaussians at the current step: $\boldsymbol{G} \gets \mathcal{D}\bigl(\boldsymbol{S},\, \mathcal{I}_{\sT}^{(t_i)},\, \mathcal{P}_{\sT},\, t_i\bigr).$
    \State Compute velocity using the rendered images: $\{\boldsymbol{v}^{j} \;|\; \boldsymbol{v}^{j} = \bigl(\mathcal{R}(\boldsymbol{G},\boldsymbol{P}^{j}_{\sT}) - {\boldsymbol{I}}^{j,(t_i)}_{\sT}\bigr) / (1 - t_i)\}_{j=1}^{V_\sT}$.
    \State Advance the target images: $\mathcal{I}_{\sT}^{(t_{i+1})} = \{{\boldsymbol{I}}^{j, (t_{i+1})}_{\sT} \;|\; {\boldsymbol{I}}^{j, (t_{i+1})}_{\sT} = {\boldsymbol{I}}^{j,(t_i)}_{\sT} + \boldsymbol{v}^j (t_{i+1} {-} t_i)\}$.
    \EndFor
    \State \Return $\boldsymbol{G}$.
\end{algorithmic}
\caption{Inference with Euler Solver.}
\label{alg:inference}
\end{algorithm}

%% file: tables/regression_priors.tex
\begin{table*}[t]
    \footnotesize
    \centering
    \caption{Geometric and generative priors and parameter counts of the compared methods. A dash indicates that no corresponding prior is used or that a separate parameter count is unavailable.}
    \setlength{\tabcolsep}{0.9pt}
    \renewcommand{\arraystretch}{1.1}
    \begin{tabular}{l cc ccc}
        \toprule
        \multirow{2}{*}{\textbf{Method}}
        & \multicolumn{2}{c}{\textbf{Prior}}
        & \multicolumn{3}{c}{\textbf{Params (B)}} \\
        \cmidrule(r{2pt}){2-3}
        \cmidrule(r{2pt}){4-6}
        & \textbf{Geometric}
        & \textbf{Generative}
        & \textbf{Encoder}
        & \textbf{Decoder}
        & \textbf{Total} \\
        \midrule
        \rowcolor{gray!20}
        \multicolumn{6}{l}{\textit{Regression-based Methods}} \\
        DepthSplat        & DA2~\citep{yang2024depthanythingv2} & -- & 0.347  & -- & 0.347 \\
        AnySplat          & VGGT~\citep{wang2025vggt}              & -- & 1.191 & -- & 1.191 \\
        YoNoSplat         & Pi3~\citep{wang2026pi3} & -- & 0.965  & -- & 0.965 \\
        DA3 & DA3~\citep{lin2025depthanythingv3} & -- & 1.356     & -- & 1.356 \\
        NoPoSplat         & MASt3R~\citep{leroy2024mast3r}            & -- & 0.612     & -- & 0.612 \\
        NAS3R             & VGGT~\citep{wang2025vggt}              & -- & 1.223     & -- & 1.223 \\
        SPFSplat          & MASt3R~\citep{leroy2024mast3r}             & -- & 0.613      & -- & 0.613 \\
        \midrule
        \rowcolor{gray!20}
        \multicolumn{6}{l}{\textit{Generative Methods}} \\
        Bolt3D          & MASt3R~\citep{leroy2024mast3r}   & CAT3D~\citep{gao2024cat3d}   & --    & --      & -- \\
        latentSplat     & --      & VAE-GAN~\citep{rombach2022sd} & 0.101    & 0.084      & 0.185 \\
        MVSplat360      & MVSplat~\citep{chen2024mvsplat} & SVD~\citep{blattmann2023svd}     & 0.012 & 2.389 & 2.401 \\
        ReconSplat      & MVSplat~\citep{chen2024mvsplat} & SD~\citep{rombach2022sd}      & 0.012    & 1.298     & 1.310 \\
        \textbf{\ours{}(ViT-L)} & --  & --      & \textbf{0.304} & \textbf{0.131} & \textbf{0.434} \\
        \textbf{\ours{}} & VGGT~\citep{wang2025vggt}  & --      & \textbf{0.912} & \textbf{0.131} & \textbf{1.042} \\
        \bottomrule
    \end{tabular}
    \label{tab:regression_priors}
    \vspace{-8pt}
\end{table*}

%% file: tables/acid.tex
\begin{table}[t]
    \centering
    \footnotesize
    \caption{Out-of-domain novel-view synthesis results on ACID.}
    \setlength{\tabcolsep}{20pt}
    \renewcommand{\arraystretch}{1.1}
    \begin{tabular}{
        l
        w{c}{0.6cm} w{c}{0.6cm} w{c}{0.6cm} w{c}{0.6cm}
    }
        \toprule
        \textbf{Method} & PSNR $\uparrow$ & SSIM $\uparrow$ & LPIPS $\downarrow$ & FID $\downarrow$ \\
        \midrule
        \rowcolor{gray!20}
        \multicolumn{5}{l}{\textit{Regression-based Methods}} \\
        MVSplat    & \secondbest{28.25} & \secondbest{0.843} & \secondbest{0.144} & \secondbest{6.38} \\
        DepthSplat & 19.65 & 0.628 & 0.298 & 28.36 \\
        AnySplat   & 21.49 & 0.631 & 0.300 & 8.47 \\
        YoNoSplat  & 25.72 & 0.768 & 0.185 & 10.14 \\
        \rowcolor{gray!20}
        \multicolumn{5}{l}{\textit{Generative Methods}} \\
        latentSplat & 25.78 & 0.789 & 0.185 & 6.55 \\
        MVSplat360  & 21.91 & 0.686 & 0.296 & 25.24 \\
        ReconSplat  & 22.53 & 0.656 & 0.286 & 17.28 \\
        \cellcolor{ourscell}\textbf{\ours{}}
        & \cellcolor{ourscell}\best{29.38}
        & \cellcolor{ourscell}\best{0.866}
        & \cellcolor{ourscell}\best{0.138}
        & \cellcolor{ourscell}\best{6.29} \\
        \bottomrule
    \end{tabular}
    \label{tab:acid}
    \vspace{-10pt}
\end{table}

%% file: tables/additional_regression_extrapolation.tex
\begin{table*}[t]
    \footnotesize
    \centering
    \caption{Comparison with additional regression-based methods on 2-view RE10K extrapolation.}
    \setlength{\tabcolsep}{14pt}
    \renewcommand{\arraystretch}{1.1}
    \begin{tabular}{lcccc}
        \toprule
        \textbf{Method}
        & PSNR $\uparrow$
        & SSIM $\uparrow$
        & LPIPS $\downarrow$
        & FID $\downarrow$ \\
        \midrule
        Depth Anything V3 & 17.91 & 0.574 & 0.339 & 13.28 \\
        NoPoSplat         & 21.41 & 0.757 & 0.207 & 6.09 \\
        NAS3R             & 19.92 & 0.782 & 0.205 & 7.78 \\
        SPFSplat          & 20.22 & 0.775 & 0.210 & 7.49 \\
        \midrule
        \cellcolor{ourscell}\textbf{\ours{}}
        & \cellcolor{ourscell}\best{26.03}
        & \cellcolor{ourscell}\best{0.872}
        & \cellcolor{ourscell}\best{0.141}
        & \cellcolor{ourscell}\best{2.75} \\
        \bottomrule
    \end{tabular}
    \label{tab:additional_regression_extrapolation}
    \vspace{-5pt}
\end{table*}

%% file: reference.bib
@String(CVPR= {IEEE Conf. Comput. Vis. Pattern Recog.})

@String(TOG= {ACM Trans. Graph.})

@String(CVPR  = {CVPR})

@String(TOG   = {ACM TOG})

@article{mildenhall2021nerf,
  title={Nerf: Representing scenes as neural radiance fields for view synthesis},
  author={Mildenhall, Ben and Srinivasan, Pratul P and Tancik, Matthew and Barron, Jonathan T and Ramamoorthi, Ravi and Ng, Ren},
  journal={Communications of the ACM},
  volume={65},
  number={1},
  pages={99--106},
  year={2021},
  publisher={ACM New York, NY, USA}
}

@article{kerbl20233dgs,
  title={3D Gaussian Splatting for Real-Time Radiance Field Rendering.},
  author={Kerbl, Bernhard and Kopanas, Georgios and Leimk{\"u}hler, Thomas and Drettakis, George},
  journal={ACM Trans. Graph.},
  volume={42},
  number={4},
  pages={139--1},
  year={2023}
}

@inproceedings{schonberger2016sfm,
  title={Structure-from-motion revisited},
  author={Schonberger, Johannes L and Frahm, Jan-Michael},
  booktitle={Proceedings of the IEEE conference on computer vision and pattern recognition},
  pages={4104--4113},
  year={2016}
}

@inproceedings{leroy2024mast3r,
  title={Grounding image matching in 3d with mast3r},
  author={Leroy, Vincent and Cabon, Yohann and Revaud, J{\'e}r{\^o}me},
  booktitle={European Conference on Computer Vision},
  pages={71--91},
  year={2024},
  organization={Springer}
}

@inproceedings{charatan2024pixelsplat,
  title={pixelsplat: 3d gaussian splats from image pairs for scalable generalizable 3d reconstruction},
  author={Charatan, David and Li, Sizhe Lester and Tagliasacchi, Andrea and Sitzmann, Vincent},
  booktitle={Proceedings of the IEEE/CVF Conference on Computer Vision and Pattern Recognition},
  pages={19457--19467},
  year={2024}
}

@inproceedings{chen2024mvsplat,
  title={Mvsplat: Efficient 3d gaussian splatting from sparse multi-view images},
  author={Chen, Yuedong and Xu, Haofei and Zheng, Chuanxia and Zhuang, Bohan and Pollefeys, Marc and Geiger, Andreas and Cham, Tat-Jen and Cai, Jianfei},
  booktitle={European Conference on Computer Vision},
  pages={370--386},
  year={2024},
  organization={Springer}
}

@inproceedings{
ye2025noposplat,
title={No Pose, No Problem: Surprisingly Simple 3D Gaussian Splats from Sparse Unposed Images},
author={Botao Ye and Sifei Liu and Haofei Xu and Xueting Li and Marc Pollefeys and Ming-Hsuan Yang and Songyou Peng},
booktitle={The Thirteenth International Conference on Learning Representations},
year={2025},
}

@inproceedings{ling2024dl3dv,
  title={Dl3dv-10k: A large-scale scene dataset for deep learning-based 3d vision},
  author={Ling, Lu and Sheng, Yichen and Tu, Zhi and Zhao, Wentian and Xin, Cheng and Wan, Kun and Yu, Lantao and Guo, Qianyu and Yu, Zixun and Lu, Yawen and others},
  booktitle={Proceedings of the IEEE/CVF Conference on Computer Vision and Pattern Recognition},
  pages={22160--22169},
  year={2024}
}

@article{wang2004ssim,
  title={Image quality assessment: from error visibility to structural similarity},
  author={Wang, Zhou and Bovik, Alan C and Sheikh, Hamid R and Simoncelli, Eero P},
  journal={IEEE transactions on image processing},
  volume={13},
  number={4},
  pages={600--612},
  year={2004},
  publisher={IEEE}
}

@inproceedings{zhang2018lpips,
  title={The unreasonable effectiveness of deep features as a perceptual metric},
  author={Zhang, Richard and Isola, Phillip and Efros, Alexei A and Shechtman, Eli and Wang, Oliver},
  booktitle={Proceedings of the IEEE conference on computer vision and pattern recognition},
  pages={586--595},
  year={2018}
}

@inproceedings{
dosovitskiy2021vit,
title={An Image is Worth 16x16 Words: Transformers for Image Recognition at Scale},
author={Alexey Dosovitskiy and Lucas Beyer and Alexander Kolesnikov and Dirk Weissenborn and Xiaohua Zhai and Thomas Unterthiner and Mostafa Dehghani and Matthias Minderer and Georg Heigold and Sylvain Gelly and Jakob Uszkoreit and Neil Houlsby},
booktitle={International Conference on Learning Representations},
year={2021},
}

@inproceedings{wang2024dust3r,
  title={Dust3r: Geometric 3d vision made easy},
  author={Wang, Shuzhe and Leroy, Vincent and Cabon, Yohann and Chidlovskii, Boris and Revaud, Jerome},
  booktitle={Proceedings of the IEEE/CVF Conference on Computer Vision and Pattern Recognition},
  pages={20697--20709},
  year={2024}
}

@article{zhou2018stereo,
  title={Stereo magnification: learning view synthesis using multiplane images},
  author={Zhou, Tinghui and Tucker, Richard and Flynn, John and Fyffe, Graham and Snavely, Noah},
  journal={ACM Transactions on Graphics (TOG)},
  volume={37},
  number={4},
  pages={1--12},
  year={2018},
  publisher={ACM New York, NY, USA}
}

@inproceedings{liu2021infinite,
  title={Infinite nature: Perpetual view generation of natural scenes from a single image},
  author={Liu, Andrew and Tucker, Richard and Jampani, Varun and Makadia, Ameesh and Snavely, Noah and Kanazawa, Angjoo},
  booktitle={Proceedings of the IEEE/CVF International Conference on Computer Vision},
  pages={14458--14467},
  year={2021}
}

@inproceedings{ronneberger2015unet,
  title={U-net: Convolutional networks for biomedical image segmentation},
  author={Ronneberger, Olaf and Fischer, Philipp and Brox, Thomas},
  booktitle={Medical image computing and computer-assisted intervention--MICCAI 2015: 18th international conference, Munich, Germany, October 5-9, 2015, proceedings, part III 18},
  pages={234--241},
  year={2015},
  organization={Springer}
}

@inproceedings{wang2025vggt,
  title={Vggt: Visual geometry grounded transformer},
  author={Wang, Jianyuan and Chen, Minghao and Karaev, Nikita and Vedaldi, Andrea and Rupprecht, Christian and Novotny, David},
  booktitle={Proceedings of the Computer Vision and Pattern Recognition Conference},
  pages={5294--5306},
  year={2025}
}

@inproceedings{huang2025spfsplat,
  title={No Pose at All: Self-Supervised Pose-Free 3D Gaussian Splatting from Sparse Views},
  author={Huang, Ranran and Mikolajczyk, Krystian},
  booktitle={Proceedings of the IEEE/CVF International Conference on Computer Vision},
  year={2025},
  pages={27947--27957}
}

@article{huang2025spfsplatv2,
  title={SPFSplatV2: Efficient Self-Supervised Pose-Free 3D Gaussian Splatting from Sparse Views},
  author={Huang, Ranran and Mikolajczyk, Krystian},
  journal={arXiv preprint arXiv:2509.17246},
  year={2025}
}

@article{jiang2025anysplat,
  title={Anysplat: Feed-forward 3d gaussian splatting from unconstrained views},
  author={Jiang, Lihan and Mao, Yucheng and Xu, Linning and Lu, Tao and Ren, Kerui and Jin, Yichen and Xu, Xudong and Yu, Mulin and Pang, Jiangmiao and Zhao, Feng and others},
  journal={ACM Transactions on Graphics (TOG)},
  volume={44},
  number={6},
  pages={1--16},
  year={2025},
  publisher={ACM, New York, NY, USA}
}

@inproceedings{xu2025depthsplat,
  title={Depthsplat: Connecting gaussian splatting and depth},
  author={Xu, Haofei and Peng, Songyou and Wang, Fangjinhua and Blum, Hermann and Barath, Daniel and Geiger, Andreas and Pollefeys, Marc},
  booktitle={Proceedings of the Computer Vision and Pattern Recognition Conference},
  pages={16453--16463},
  year={2025}
}

@inproceedings{huang2026nas3r,
    author    = {Huang, Ranran and Luo, Weixun and Mao, Ye and Mikolajczyk, Krystian},
    title     = {From None to All: Self-Supervised 3D Reconstruction via Novel View Synthesis},
    booktitle = {Proceedings of the IEEE/CVF Conference on Computer Vision and Pattern Recognition (CVPR)},
    year      = {2026},
    pages     = {37358-37369}
}

@inproceedings{zhao2026erayzer,
  title={E-rayzer: Self-supervised 3d reconstruction as spatial visual pre-training},
  author={Zhao, Qitao and Tan, Hao and Wang, Qianqian and Bi, Sai and Zhang, Kai and Sunkavalli, Kalyan and Tulsiani, Shubham and Jiang, Hanwen},
  booktitle={Proceedings of the IEEE/CVF Conference on Computer Vision and Pattern Recognition},
  pages={7525--7535},
  year={2026}
}

@inproceedings{sargent2024zeronvs,
  title={Zeronvs: Zero-shot 360-degree view synthesis from a single image},
  author={Sargent, Kyle and Li, Zizhang and Shah, Tanmay and Herrmann, Charles and Yu, Hong-Xing and Zhang, Yunzhi and Chan, Eric Ryan and Lagun, Dmitry and Fei-Fei, Li and Sun, Deqing and others},
  booktitle={Proceedings of the IEEE/CVF Conference on Computer Vision and Pattern Recognition},
  pages={9420--9429},
  year={2024}
}

@inproceedings{liu2023zero123,
  title={Zero-1-to-3: Zero-shot one image to 3d object},
  author={Liu, Ruoshi and Wu, Rundi and Van Hoorick, Basile and Tokmakov, Pavel and Zakharov, Sergey and Vondrick, Carl},
  booktitle={Proceedings of the IEEE/CVF international conference on computer vision},
  pages={9298--9309},
  year={2023}
}

@article{nichol2022pointe,
  title={Point-e: A system for generating 3d point clouds from complex prompts},
  author={Nichol, Alex and Jun, Heewoo and Dhariwal, Prafulla and Mishkin, Pamela and Chen, Mark},
  journal={arXiv preprint arXiv:2212.08751},
  year={2022}
}

@inproceedings{xiang2025trellis,
  title={Structured 3d latents for scalable and versatile 3d generation},
  author={Xiang, Jianfeng and Lv, Zelong and Xu, Sicheng and Deng, Yu and Wang, Ruicheng and Zhang, Bowen and Chen, Dong and Tong, Xin and Yang, Jiaolong},
  booktitle={Proceedings of the IEEE/CVF conference on computer vision and pattern recognition},
  pages={21469--21480},
  year={2025}
}

@inproceedings{anciukevivcius2024denoising,
  title={Denoising diffusion via image-based rendering},
  author={Anciukevicius, Titas and Manhardt, Fabian and Tombari, Federico and Henderson, Paul},
  booktitle={The Twelfth International Conference on Learning Representations},
  year={2024}
}

@inproceedings{anciukevivcius2023renderdiffusion,
  title={Renderdiffusion: Image diffusion for 3d reconstruction, inpainting and generation},
  author={Anciukevicius, Titas and Xu, Zexiang and Fisher, Matthew and Henderson, Paul and Bilen, Hakan and Mitra, Niloy J and Guerrero, Paul},
  booktitle={Proceedings of the IEEE/CVF conference on computer vision and pattern recognition},
  pages={12608--12618},
  year={2023}
}

@inproceedings{xu2024dmv3d,
  title={Dmv3d: Denoising multi-view diffusion using 3d large reconstruction model},
  author={Xu, Yinghao and Tan, Hao and Luan, Fujun and Bi, Sai and Wang, Peng and Li, Jiahao and Shi, Zifan and Sunkavalli, Kalyan and Wetzstein, Gordon and Xu, Zexiang and others},
  booktitle={International Conference on Learning Representations},
  volume={2024},
  pages={33168--33184},
  year={2024}
}

@inproceedings{lin2025diffsplat,
  title={Diffsplat: Repurposing image diffusion models for scalable gaussian splat generation},
  author={Lin, Chenguo and Pan, Panwang and Yang, Bangbang and Li, Zeming and Mu, Yadong},
  booktitle={International Conference on Learning Representations},
  volume={2025},
  pages={40745--40767},
  year={2025}
}

@inproceedings{hollein2024viewdiff,
  title={Viewdiff: 3d-consistent image generation with text-to-image models},
  author={Hollein, Lukas and Bozic, Aljaz and Muller, Norman and Novotny, David and Tseng, Hung-Yu and Richardt, Christian and Zollhofer, Michael and Niessner, Matthias},
  booktitle={Proceedings of the IEEE/CVF conference on computer vision and pattern recognition},
  pages={5043--5052},
  year={2024}
}

@inproceedings{
shi2024mvdream,
title={{MVD}ream: Multi-view Diffusion for 3D Generation},
author={Yichun Shi and Peng Wang and Jianglong Ye and Long Mai and Kejie Li and Xiao Yang},
booktitle={The Twelfth International Conference on Learning Representations},
year={2024},
}

@inproceedings{chan2023genvs,
  title={Generative novel view synthesis with 3d-aware diffusion models},
  author={Chan, Eric R and Nagano, Koki and Chan, Matthew A and Bergman, Alexander W and Park, Jeong Joon and Levy, Axel and Aittala, Miika and De Mello, Shalini and Karras, Tero and Wetzstein, Gordon},
  booktitle={Proceedings of the IEEE/CVF International Conference on Computer Vision},
  pages={4217--4229},
  year={2023}
}

@inproceedings{wu2024reconfusion,
  title={Reconfusion: 3d reconstruction with diffusion priors},
  author={Wu, Rundi and Mildenhall, Ben and Henzler, Philipp and Park, Keunhong and Gao, Ruiqi and Watson, Daniel and Srinivasan, Pratul P and Verbin, Dor and Barron, Jonathan T and Poole, Ben and others},
  booktitle={Proceedings of the IEEE/CVF conference on computer vision and pattern recognition},
  pages={21551--21561},
  year={2024}
}

@article{gao2024cat3d,
  title={Cat3d: Create anything in 3d with multi-view diffusion models},
  author={Gao, Ruiqi and Holynski, Aleksander and Henzler, Philipp and Brussee, Arthur and Martin-Brualla, Ricardo and Srinivasan, Pratul and Barron, Jonathan T and Poole, Ben},
  journal={arXiv preprint arXiv:2405.10314},
  year={2024}
}

@inproceedings{long2024wonder3d,
  title={Wonder3d: Single image to 3d using cross-domain diffusion},
  author={Long, Xiaoxiao and Guo, Yuan-Chen and Lin, Cheng and Liu, Yuan and Dou, Zhiyang and Liu, Lingjie and Ma, Yuexin and Zhang, Song-Hai and Habermann, Marc and Theobalt, Christian and others},
  booktitle={Proceedings of the IEEE/CVF conference on computer vision and pattern recognition},
  pages={9970--9980},
  year={2024}
}

@article{chen2024mvsplat360,
  title={Mvsplat360: Feed-forward 360 scene synthesis from sparse views},
  author={Chen, Yuedong and Zheng, Chuanxia and Xu, Haofei and Zhuang, Bohan and Vedaldi, Andrea and Cham, Tat-Jen and Cai, Jianfei},
  journal={Advances in Neural Information Processing Systems},
  volume={37},
  pages={107064--107086},
  year={2024}
}

@inproceedings{
he2025cameractrl,
title={CameraCtrl: Enabling Camera Control for Video Diffusion Models},
author={Hao He and Yinghao Xu and Yuwei Guo and Gordon Wetzstein and Bo Dai and Hongsheng Li and Ceyuan Yang},
booktitle={The Thirteenth International Conference on Learning Representations},
year={2025},
}

@article{liu2026reconx,
  title={Reconx: Reconstruct any scene from sparse views with video diffusion model},
  author={Liu, Fangfu and Sun, Wenqiang and Wang, Hanyang and Wang, Yikai and Sun, Haowen and Ye, Junliang and Zhang, Jun and Duan, Yueqi},
  journal={IEEE Transactions on Image Processing},
  year={2026},
  publisher={IEEE}
}

@inproceedings{sun2025dimensionx,
  title={Dimensionx: Create any 3d and 4d scenes from a single image with decoupled video diffusion},
  author={Sun, Wenqiang and Chen, Shuo and Liu, Fangfu and Chen, Zilong and Duan, Yueqi and Zhu, Jun and Zhang, Jun and Wang, Yikai},
  booktitle={Proceedings of the IEEE/CVF International Conference on Computer Vision},
  pages={13695--13706},
  year={2025}
}

@inproceedings{yu2025wonderworld,
  title={Wonderworld: Interactive 3d scene generation from a single image},
  author={Yu, Hong-Xing and Duan, Haoyi and Herrmann, Charles and Freeman, William T and Wu, Jiajun},
  booktitle={Proceedings of the Computer Vision and Pattern Recognition Conference},
  pages={5916--5926},
  year={2025}
}

@article{yu2025viewcrafter,
  title={ViewCrafter: Taming Video Diffusion Models for High-fidelity Novel View Synthesis},
  author={Yu, Wangbo and Xing, Jinbo and Yuan, Li and Hu, Wenbo and Li, Xiaoyu and Huang, Zhipeng and Gao, Xiangjun and Wong, Tien-Tsin and Shan, Ying and Tian, Yonghong},
  journal={IEEE Transactions on Pattern Analysis and Machine Intelligence},
  year={2025},
  publisher={IEEE}
}

@inproceedings{szymanowicz2025bolt3d,
  title={Bolt3d: Generating 3d scenes in seconds},
  author={Szymanowicz, Stanislaw and Zhang, Jason Y and Srinivasan, Pratul and Gao, Ruiqi and Brussee, Arthur and Holynski, Aleksander and Martin-Brualla, Ricardo and Barron, Jonathan T and Henzler, Philipp},
  booktitle={Proceedings of the IEEE/CVF International Conference on Computer Vision},
  pages={24846--24857},
  year={2025}
}

@inproceedings{wewer2024latentsplat,
  title={latentsplat: Autoencoding variational gaussians for fast generalizable 3d reconstruction},
  author={Wewer, Christopher and Raj, Kevin and Ilg, Eddy and Schiele, Bernt and Lenssen, Jan Eric},
  booktitle={European conference on computer vision},
  pages={456--473},
  year={2024},
  organization={Springer}
}

@inproceedings{cai2025diffusiongs,
  title={Baking gaussian splatting into diffusion denoiser for fast and scalable single-stage image-to-3d generation and reconstruction},
  author={Cai, Yuanhao and Zhang, He and Zhang, Kai and Liang, Yixun and Ren, Mengwei and Luan, Fujun and Liu, Qing and Kim, Soo Ye and Zhang, Jianming and Zhang, Zhifei and others},
  booktitle={Proceedings of the IEEE/CVF International Conference on Computer Vision},
  pages={25062--25072},
  year={2025}
}

@inproceedings{zhang2024gslrm,
  title={Gs-lrm: Large reconstruction model for 3d gaussian splatting},
  author={Zhang, Kai and Bi, Sai and Tan, Hao and Xiangli, Yuanbo and Zhao, Nanxuan and Sunkavalli, Kalyan and Xu, Zexiang},
  booktitle={European Conference on Computer Vision},
  pages={1--19},
  year={2024},
  organization={Springer}
}

@inproceedings{
hong2024lrm,
title={{LRM}: Large Reconstruction Model for Single Image to 3D},
author={Yicong Hong and Kai Zhang and Jiuxiang Gu and Sai Bi and Yang Zhou and Difan Liu and Feng Liu and Kalyan Sunkavalli and Trung Bui and Hao Tan},
booktitle={The Twelfth International Conference on Learning Representations},
year={2024},
}

@inproceedings{xu2024grm,
  title={Grm: Large gaussian reconstruction model for efficient 3d reconstruction and generation},
  author={Xu, Yinghao and Shi, Zifan and Yifan, Wang and Chen, Hansheng and Yang, Ceyuan and Peng, Sida and Shen, Yujun and Wetzstein, Gordon},
  booktitle={European Conference on Computer Vision},
  pages={1--20},
  year={2024},
  organization={Springer}
}

@inproceedings{tang2024lgm,
  title={Lgm: Large multi-view gaussian model for high-resolution 3d content creation},
  author={Tang, Jiaxiang and Chen, Zhaoxi and Chen, Xiaokang and Wang, Tengfei and Zeng, Gang and Liu, Ziwei},
  booktitle={European Conference on Computer Vision},
  pages={1--18},
  year={2024},
  organization={Springer}
}

@inproceedings{muller2023diffrf,
  title={Diffrf: Rendering-guided 3d radiance field diffusion},
  author={M{\"u}ller, Norman and Siddiqui, Yawar and Porzi, Lorenzo and Bulo, Samuel Rota and Kontschieder, Peter and Nie{\ss}ner, Matthias},
  booktitle={Proceedings of the IEEE/CVF conference on computer vision and pattern recognition},
  pages={4328--4338},
  year={2023}
}

@inproceedings{shue20233triplane,
  title={3d neural field generation using triplane diffusion},
  author={Shue, J Ryan and Chan, Eric Ryan and Po, Ryan and Ankner, Zachary and Wu, Jiajun and Wetzstein, Gordon},
  booktitle={Proceedings of the IEEE/CVF conference on computer vision and pattern recognition},
  pages={20875--20886},
  year={2023}
}

@inproceedings{
zhang2024gaussiancube,
title={GaussianCube: A Structured and Explicit Radiance Representation for 3D Generative Modeling},
author={Bowen Zhang and Yiji Cheng and Jiaolong Yang and Chunyu Wang and Feng Zhao and Yansong Tang and Dong Chen and Baining Guo},
booktitle={The Thirty-eighth Annual Conference on Neural Information Processing Systems},
year={2024},
}

@inproceedings{
lipman2023flow,
title={Flow Matching for Generative Modeling},
author={Yaron Lipman and Ricky T. Q. Chen and Heli Ben-Hamu and Maximilian Nickel and Matthew Le},
booktitle={The Eleventh International Conference on Learning Representations },
year={2023},
}

@inproceedings{
liu2023flow,
title={Flow Straight and Fast: Learning to Generate and Transfer Data with Rectified Flow},
author={Xingchao Liu and Chengyue Gong and qiang liu},
booktitle={The Eleventh International Conference on Learning Representations },
year={2023},
}

@inproceedings{li2026jit,
  title={Back to basics: Let denoising generative models denoise},
  author={Li, Tianhong and He, Kaiming},
  booktitle={Proceedings of the IEEE/CVF Conference on Computer Vision and Pattern Recognition},
  pages={36115--36125},
  year={2026}
}

@article{ho2020ddpm,
  title={Denoising diffusion probabilistic models},
  author={Ho, Jonathan and Jain, Ajay and Abbeel, Pieter},
  journal={Advances in neural information processing systems},
  volume={33},
  pages={6840--6851},
  year={2020}
}

@inproceedings{
song2021ddim,
title={Denoising Diffusion Implicit Models},
author={Jiaming Song and Chenlin Meng and Stefano Ermon},
booktitle={International Conference on Learning Representations},
year={2021},
}

@inproceedings{
albergo2023buildingflow,
title={Building Normalizing Flows with Stochastic Interpolants},
author={Michael Samuel Albergo and Eric Vanden-Eijnden},
booktitle={The Eleventh International Conference on Learning Representations },
year={2023},
}

@inproceedings{rombach2022stablediffusion,
  title={High-resolution image synthesis with latent diffusion models},
  author={Rombach, Robin and Blattmann, Andreas and Lorenz, Dominik and Esser, Patrick and Ommer, Bj{\"o}rn},
  booktitle={Proceedings of the IEEE/CVF conference on computer vision and pattern recognition},
  pages={10684--10695},
  year={2022}
}

@inproceedings{peebles2023dit,
  title={Scalable diffusion models with transformers},
  author={Peebles, William and Xie, Saining},
  booktitle={Proceedings of the IEEE/CVF international conference on computer vision},
  pages={4195--4205},
  year={2023}
}

@inproceedings{ma2024sit,
  title={Sit: Exploring flow and diffusion-based generative models with scalable interpolant transformers},
  author={Ma, Nanye and Goldstein, Mark and Albergo, Michael S and Boffi, Nicholas M and Vanden-Eijnden, Eric and Xie, Saining},
  booktitle={European Conference on Computer Vision},
  pages={23--40},
  year={2024},
  organization={Springer}
}

@inproceedings{
song2021scorebased,
title={Score-Based Generative Modeling through Stochastic Differential Equations},
author={Yang Song and Jascha Sohl-Dickstein and Diederik P Kingma and Abhishek Kumar and Stefano Ermon and Ben Poole},
booktitle={International Conference on Learning Representations},
year={2021},
}

@inproceedings{nichol2021improved,
  title={Improved denoising diffusion probabilistic models},
  author={Nichol, Alexander Quinn and Dhariwal, Prafulla},
  booktitle={International conference on machine learning},
  pages={8162--8171},
  year={2021},
  organization={PMLR}
}

@article{dhariwal2021diffusion,
  title={Diffusion models beat gans on image synthesis},
  author={Dhariwal, Prafulla and Nichol, Alexander},
  journal={Advances in neural information processing systems},
  volume={34},
  pages={8780--8794},
  year={2021}
}

@article{chen2025pixelflow,
  title={Pixelflow: Pixel-space generative models with flow},
  author={Chen, Shoufa and Ge, Chongjian and Zhang, Shilong and Sun, Peize and Luo, Ping},
  journal={arXiv preprint arXiv:2504.07963},
  year={2025}
}

@inproceedings{yu2026pixeldit,
  title={Pixeldit: Pixel diffusion transformers for image generation},
  author={Yu, Yongsheng and Xiong, Wei and Nie, Weili and Sheng, Yichen and Liu, Shiqiu and Luo, Jiebo},
  booktitle={Proceedings of the IEEE/CVF Conference on Computer Vision and Pattern Recognition},
  pages={14273--14282},
  year={2026}
}

@inproceedings{chen2026dip,
  title={Dip: Taming diffusion models in pixel space},
  author={Chen, Zhennan and Zhu, Junwei and Chen, Xu and Zhang, Jiangning and Hu, Xiaobin and Zhao, Hanzhen and Wang, Chengjie and Yang, Jian and Tai, Ying},
  booktitle={Proceedings of the IEEE/CVF Conference on Computer Vision and Pattern Recognition},
  pages={36136--36146},
  year={2026}
}

@article{lu2026onestep,
  title={One-step Latent-free Image Generation with Pixel Mean Flows},
  author={Lu, Yiyang and Lu, Susie and Sun, Qiao and Zhao, Hanhong and Jiang, Zhicheng and Wang, Xianbang and Li, Tianhong and Geng, Zhengyang and He, Kaiming},
  journal={arXiv preprint arXiv:2601.22158},
  year={2026}
}

@article{zhu2005semi,
  title={Semi-supervised learning literature survey},
  author={Zhu, Xiaojin Jerry},
  year={2005},
  publisher={University of Wisconsin-Madison Department of Computer Sciences},
  journal={Computer Sciences TR}
}

@inproceedings{szymanowicz2026lagernvs,
  title={LagerNVS: Latent geometry for fully neural real-time novel view synthesis},
  author={Szymanowicz, Stanislaw and Chen, Minghao and Wang, Jianyuan and Rupprecht, Christian and Vedaldi, Andrea},
  booktitle={Proceedings of the IEEE/CVF Conference on Computer Vision and Pattern Recognition},
  pages={15443--15453},
  year={2026}
}

@article{ba2016layernorm,
  title={Layer normalization},
  author={Ba, Jimmy Lei and Kiros, Jamie Ryan and Hinton, Geoffrey E},
  journal={arXiv preprint arXiv:1607.06450},
  year={2016}
}

@article{zhang2019rmsnorm,
  title={Root mean square layer normalization},
  author={Zhang, Biao and Sennrich, Rico},
  journal={Advances in neural information processing systems},
  volume={32},
  year={2019}
}

@article{su2024roformer,
  title={Roformer: Enhanced transformer with rotary position embedding},
  author={Su, Jianlin and Ahmed, Murtadha and Lu, Yu and Pan, Shengfeng and Bo, Wen and Liu, Yunfeng},
  journal={Neurocomputing},
  volume={568},
  pages={127063},
  year={2024},
  publisher={Elsevier}
}

@article{heusel2017fid,
  title={Gans trained by a two time-scale update rule converge to a local nash equilibrium},
  author={Heusel, Martin and Ramsauer, Hubert and Unterthiner, Thomas and Nessler, Bernhard and Hochreiter, Sepp},
  journal={Advances in neural information processing systems},
  volume={30},
  year={2017}
}

@inproceedings{ye2026yonosplat,
  title={Yonosplat: You only need one model for feedforward 3d gaussian splatting},
  author={Ye, Botao and Chen, Boqi and Xu, Haofei and Barath, Daniel and Pollefeys, Marc},
  booktitle={International Conference on Learning Representations},
  volume={2026},
  pages={39852--39871},
  year={2026}
}

@article{blattmann2023svd,
  title={Stable video diffusion: Scaling latent video diffusion models to large datasets},
  author={Blattmann, Andreas and Dockhorn, Tim and Kulal, Sumith and Mendelevitch, Daniel and Kilian, Maciej and Lorenz, Dominik and Levi, Yam and English, Zion and Voleti, Vikram and Letts, Adam and others},
  journal={arXiv preprint arXiv:2311.15127},
  year={2023}
}

@inproceedings{rombach2022sd,
  title={High-resolution image synthesis with latent diffusion models},
  author={Rombach, Robin and Blattmann, Andreas and Lorenz, Dominik and Esser, Patrick and Ommer, Bj{\"o}rn},
  booktitle={2022 IEEE/CVF conference on computer vision and pattern recognition (CVPR)},
  pages={10674--10685},
  year={2022},
  organization={ieee}
}

@article{wan2025wan,
  title={Wan: Open and advanced large-scale video generative models},
  author={Wan, Team and Wang, Ang and Ai, Baole and Wen, Bin and Mao, Chaojie and Xie, Chen-Wei and Chen, Di and Yu, Feiwu and Zhao, Haiming and Yang, Jianxiao and others},
  journal={arXiv preprint arXiv:2503.20314},
  year={2025}
}

@inproceedings{wu2025difix3d+,
  title={Difix3d+: Improving 3d reconstructions with single-step diffusion models},
  author={Wu, Jay Zhangjie and Zhang, Yuxuan and Turki, Haithem and Ren, Xuanchi and Gao, Jun and Shou, Mike Zheng and Fidler, Sanja and Gojcic, Zan and Ling, Huan},
  booktitle={2025 IEEE/CVF Conference on Computer Vision and Pattern Recognition (CVPR)},
  pages={26024--26035},
  year={2025},
  organization={IEEE}
}

@inproceedings{stracquadanio2026reconsplat,
  title={ReconSplat: Generalizable 3D Scene Reconstruction Beyond Observed Views},
  author={Stracquadanio, Giuseppe and Raj, Kevin and Grabinski, Julia and Roth, Stefan},
  booktitle={European Conference on Computer Vision},
  pages={304--325},
  year={2026},
  organization={Springer}
}

@inproceedings{wang2026pi3,
  title={{$\pi^3$}: Permutation-Equivariant Visual Geometry Learning},
  author={Wang, Yifan and Zhou, Jianjun and Zhu, Haoyi and Chang, Wenzheng and Zhou, Yang and Li, Zizun and Chen, Junyi and Pang, Jiangmiao and Shen, Chunhua and He, Tong},
  booktitle={International Conference on Learning Representations},
  volume={2026},
  pages={10481--10497},
  year={2026}
}

@article{yang2024depthanythingv2,
  title={Depth anything v2},
  author={Yang, Lihe and Kang, Bingyi and Huang, Zilong and Zhao, Zhen and Xu, Xiaogang and Feng, Jiashi and Zhao, Hengshuang},
  journal={Advances in neural information processing systems},
  volume={37},
  pages={21875--21911},
  year={2024}
}

@article{lin2025depthanythingv3,
  title={Depth anything 3: Recovering the visual space from any views},
  author={Lin, Haotong and Chen, Sili and Liew, Junhao and Chen, Donny Y and Li, Zhenyu and Shi, Guang and Feng, Jiashi and Kang, Bingyi},
  journal={arXiv preprint arXiv:2511.10647},
  year={2025}
}
